\pdfoutput=1

\documentclass[11pt]{article}

\usepackage{emnlp2021}

\usepackage{times}
\usepackage{latexsym}

\usepackage[T1]{fontenc}

\usepackage[utf8]{inputenc}
\usepackage{todonotes}

\usepackage{microtype}

\usepackage[flushleft]{threeparttable}
\usepackage{todonotes}
\usepackage{enumitem}
\usepackage{algorithm}
\usepackage{algpseudocode}
\usepackage{booktabs}
\usepackage{subcaption}
\usepackage[nointegrals]{wasysym}
\usepackage{tikz}
\usepackage{tabu}
\usepackage{tabularx}
\usepackage{float}
\usepackage{listings}
\usepackage{color}
\usepackage{multirow}
\usepackage{colortbl}
\usepackage{framed}
\usepackage{pifont}
\usepackage{dsfont}
\usepackage{xcolor}
\usepackage{amsmath,amsthm}
\usepackage{amssymb}
\usepackage{graphicx}
\usepackage{textcomp}
\usepackage{tcolorbox}
\usepackage{soul}
\usepackage{todonotes}
\usepackage{lipsum}
\definecolor{DarkRed}{RGB}{130,25,0}

\newcommand{\tbd}{\textrm{TB-Dense}}

\newcommand{\bert}{BERT$_{LARGE}$}

\newcommand{\roberta}{RoBERTa$_{LARGE}$}

\newcommand{\econet}{\emph{ECONET}}
\newcommand{\torque}{\textsc{TORQUE}}
\newcommand{\matres}{\textsc{Matres}}
\newcommand{\red}{\textsc{RED}}
\newcommand{\mctaco}{\textsc{MCTACO}}

\newcommand{\ignore}[1]{}
\definecolor{mypurple}{RGB}{121,55,196}
\definecolor{myblue}{RGB}{60,177,245}
\definecolor{myorange}{RGB}{243,206,84}
\definecolor{mygreen}{RGB}{41,222,35}
\definecolor{light-gray}{gray}{0.9}

\newcounter{exctr}

\newcounter{eventCtr}

\newcounter{timexCtr}

\newcommand{\temprel}[1]{\MakeUppercase{\textit{#1}}}

\title{ECONET: Effective Continual Pretraining of Language Models for Event Temporal Reasoning}

\author{Rujun Han$^{1}$ ~ Xiang Ren$^{1}$ ~ Nanyun Peng$^{1,2}$\\
$^1$University of Southern California ~ $^2$University of California, Los Angeles \\
{\tt \{rujunhan,xiangren\}@usc.edu;violetpeng@cs.ucla.edu}
}

\begin{document}
\maketitle

\begin{abstract}
While pre-trained language models (PTLMs) have achieved noticeable success on many NLP tasks, they still struggle for tasks that require event temporal reasoning, which is essential for event-centric applications. We present a continual pre-training approach that equips PTLMs with targeted knowledge about event temporal relations. We design self-supervised learning objectives to recover masked-out event and temporal indicators and to discriminate sentences from their corrupted counterparts (where event or temporal indicators got replaced). By further pre-training a PTLM with these objectives jointly, we reinforce its attention to event and temporal information, yielding enhanced capability on event temporal reasoning. This \textbf{E}ffective \textbf{CON}tinual pre-training framework for \textbf{E}vent \textbf{T}emporal reasoning ({\econet}) improves the PTLMs' fine-tuning performances across five relation extraction and question answering tasks and achieves new or on-par state-of-the-art performances in most of our downstream tasks.\footnote{Reproduction code, training data and models are available here: https://github.com/PlusLabNLP/ECONET.}
\end{abstract}

\section{Introduction}
\label{sec:intro}
\vspace{-0.2cm}

\begin{figure}[t]
    \centering
\includegraphics[trim=0.1cm 2.7cm 0.1cm 2.7cm, clip, angle=-90, width=0.95\columnwidth]{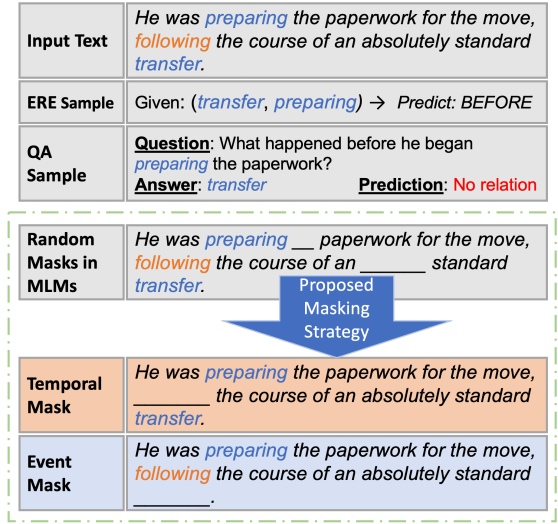}
\vspace{-0.2cm}
\caption{\textbf{Top}: an example illustrating the difference between ERE and QA / MRC samples of event temporal reasoning. \textbf{Bottom}: our targeted masking strategy for {\econet} v.s. random masking in PTLMs.}
\vspace{-0.8cm}
\label{fig:illustrating-exp}
\end{figure}

Reasoning event temporal relations is crucial for natural language understanding, and facilitates many real-world applications, such as tracking biomedical histories~\cite{rumshisky-etal-2013-eval, bethard-etal-2015-semeval, bethard-etal-2016-semeval, bethard-etal-2017-semeval}, generating stories~\cite{yao2019plan, goldfarb2020content}, and forecasting social events~\cite{li-etal-2020-connecting, jin2020forecastqa}.
In this work, we study two prominent event temporal reasoning tasks as shown in Figure~\ref{fig:illustrating-exp}: event relation extraction (ERE) \cite{ChambersTBS2014, NingWuRo18, ogorman-etal-2016-richer, MostafazadehGCAL2016} that predicts temporal relations between a pair of events, and machine reading comprehension (MRC) \citep{ning-etal-2020-torque, zhou-etal-2019-going} where a passage and a question about event temporal relations is presented, and models need to provide correct answers using the information in a given passage.

Recent approaches leveraging large pre-trained language models (PTLMs) achieved state-of-the-art results on a range of event temporal reasoning tasks \citep{ning-etal-2020-torque, pereira-etal-2020-adversarial, wang-etal-2020-joint, clinicalTemporal-zhou, contextualTemporal-han}. 
Despite the progress, vanilla PTLMs do not focus on capturing \textbf{event temporal knowledge} that can be used to infer event relations. For example, in Figure~\ref{fig:illustrating-exp}, an annotator of the QA sample can easily infer from the temporal indicator ``\textit{following}'' that ``\textit{transfer}'' happens \temprel{before} ``\textit{preparing the paperwork}'', but a fine-tuned RoBERTa model predicts that ``\textit{transfer}'' has no such relation with the event ``\textit{preparing the paperwork.}'' Plenty of such cases exist in our error analysis on PTLMs for event temporal relation-related tasks. We hypothesize that such deficiency is caused by original PTLMs' random masks in the pre-training where temporal indicators and event triggers are under-weighted and hence not attended well enough for our downstream tasks. TacoLM \citep{ZNKR20} explored the idea of targeted masking and predicting textual cues of event frequency, duration and typical time, which showed improvements over vanilla PTLMs on related tasks. However, event frequency, duration and time do not directly help machines understand pairwise event temporal relations. Moreover, the mask prediction loss of TacoLM leverages a soft cross-entropy objective, which is manually calibrated with external knowledge and could inadvertently introduce noise in the continual pre-training.

We propose {\econet}, a continual pre-training framework combining mask prediction and contrastive loss using our masked samples. Our targeted masking strategy focuses only on event triggers and temporal indicators as shown in Figure~\ref{fig:illustrating-exp}. This design assists models to concentrate on events and temporal cues, and potentially strengthen models' ability to understand event temporal relations better in the downstream tasks. We further pre-train PTLMs with the following objectives jointly: the mask prediction objective trains a generator that recovers the masked temporal indicators or events, and the contrastive loss trains a discriminator that shares the representations with the generator and determines whether a predicted masked token is corrupted or original \cite{Clark2020ELECTRA}. Our experiments demonstrate that {\econet} is effective at improving the original PTLMs' performances on event temporal reasoning.

\begin{figure*}[t]
    \centering
\includegraphics[trim=5cm 0cm 5cm 0cm, clip, width=0.75\columnwidth, angle=-90]{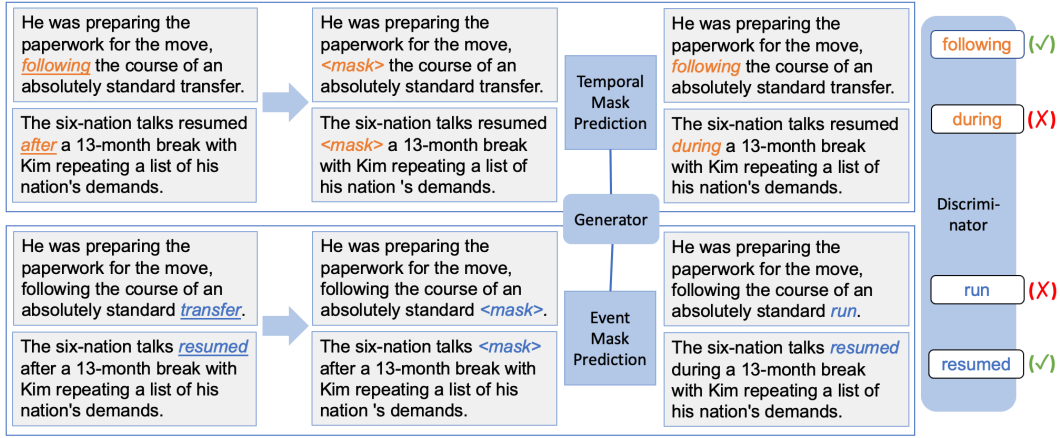}
\vspace{-0.1cm}
\caption{The proposed generator-discriminator (\econet) architecture for event temporal reasoning. The upper block is the mask prediction task for temporal indicators and the bottom block is the mask prediction task for events. Both generators and the discriminator share the same representations.}
\vspace{-0.5cm}
\label{fig:method}
\end{figure*}

\begin{table}[t]
    \small 
 	\centering
 	\setlength\extrarowheight{1pt}
 	\begin{tabular}{ll} \hline
 	\toprule
 	\textbf{Category} & Words \\
 	\toprule
 	\rowcolor{light-gray}  & before, until, previous to, prior to, \\ 
 	\rowcolor{light-gray} \multirow{-2}{*}{\textbf{[before]}} & preceding, followed by  \\
 	
 	\multirow{2}{*}{\textbf{[after]}} & after, following, since, now that\\
 	& soon after, once$^{**}$ \\
 	
 	\rowcolor{light-gray} & during, while, when, at the time, \\
 	\rowcolor{light-gray} \multirow{-2}{*}{\textbf{[during]}} & at the same time, meanwhile \\
 	
 	\multirow{2}{*}{\textbf{[past]}} & earlier, previously, formerly,\\
 	& yesterday,  in the past,  last time \\
 	
 	\rowcolor{light-gray}  & consequently, subsequently, in turn, \\
 	\rowcolor{light-gray} \multirow{-2}{*}{\textbf{[future]}} & henceforth, later, then \\
 	
 	\multirow{2}{*}{\textbf{[beginning]}} & initially, originally, at the beginning \\
 	& to begin, starting with, to start with \\
 	
 	 \rowcolor{light-gray} \textbf{[ending]} & finally, in the end, at last, lastly \\
   \bottomrule
 	\end{tabular}
   	\caption{The full list of the temporal lexicon. Categories are created based on authors' domain knowledge and best judgment. $^{**}$ `once' can be also placed into \textbf{[past]} category due to its second meaning of `previously', which we exclude to keep words unique.}
   	\label{tab:temporal-indicators}
   	\vspace{-0.5cm}
 \end{table}
 
 We briefly summarize our contributions. 1) We propose {\econet}, a novel continual pre-training framework that integrates targeted masking and contrastive loss for event temporal reasoning. 2) Our training objectives effectively learn from the targeted masked samples and inject richer event temporal knowledge in PTLMs, which leads to stronger fine-tuning performances over five widely used event temporal commonsense tasks. In most target tasks, {\econet} achieves SOTA results in comparison with existing methods. 3) Compared with full-scale pre-training, {\econet} requires a much smaller amount of training data and can cope with various PTLMs such as BERT and RoBERTa. 4) In-depth analysis shows that {\econet} successfully transfers knowledge in terms of textual cues of event triggers and relations into the target tasks, particularly under low-resource settings.

 \vspace{-0.1cm}
\section{Method}
\label{sec:method}
\vspace{-0.2cm}
Our proposed method aims at addressing the issue in vanilla PTLMs that event triggers and temporal indicators are not adequately attended for our downstream event reasoning tasks. To achieve this goal, we propose to replace the random masking in PTLMs with a targeted masking strategy designed specifically for event triggers and temporal indicators. We also propose a continual pre-training method with mask prediction and contrastive loss that allows models to effectively learn from the targeted masked samples. The benefits of our method are manifested by stronger fine-tuning performances over downstream ERE and MRC tasks. 

Our overall approach {\econet} consists of three components. 1) Creating targeted self-supervised training data by masking out temporal indicators and event triggers in the input texts; 2) leveraging mask prediction and contrastive loss to continually train PTLMs, which produces an event temporal knowledge aware language model; 3) fine-tuning the enhanced language model on downstream ERE and MRC datasets. We will discuss each of these components in the following subsections.
\vspace{-0.2cm}
\subsection{Pre-trained Masked Language Models}
The current PTLMs such as BERT \citep{BERT2018} and RoBERTa \citep{ROBERTA-19} follow a random masking strategy. Figure~\ref{fig:illustrating-exp} shows such an example where random tokens / words are masked from the input sentences. More formally, let $\boldsymbol{x} = [x_1, ..., x_n]$ be a sequence of input tokens and $x^m_t \in \boldsymbol{x}^m$ represents random masked tokens. The per-sample pre-training objective is to predict the identity ($x_t$) of $x^m_t$ with a cross-entropy loss,
\vspace{-.2cm}
\begin{align} \label{eq:Obj-mlm}
\small
\mathcal{L}_{MLM} = -\sum_{x^m_t \in \boldsymbol{x}^m} \mathcal{I}[x^m_t = x_t] \log(p(x^m_t|\boldsymbol{x}))
\end{align}
Next, we will discuss the design and creation of targeted masks, training objectives and fine-tuning approaches for different tasks.
\vspace{-0.2cm}
\subsection{Targeted Masks Creation}

\paragraph{Temporal Masks.} We first compile a lexicon of 40 common temporal indicators listed in the Table~\ref{tab:temporal-indicators} based on previous error analysis and expert knowledge in the target tasks. Those indicators in the \textbf{[before]}, \textbf{[after]} and \textbf{[during]} categories can be used to represent the most common temporal relations between events. The associated words in each of these categories are synonyms of each other. Temporal indicators in the \textbf{[past]}, \textbf{[future]}, \textbf{[beginning]} and \textbf{[ending]} categories probably do not represent pairwise event relations directly, but predicting these masked tokens may still be helpful for models to understand time anchors and hence facilitates temporal reasoning.

With the temporal lexicon, we conduct string matches over the 20-year's New York Times news articles \footnote{NYT news articles are public from 1987-2007.} and obtain over 10 million 1-2 sentence passages that contain at least 1 temporal indicators. Finally, we replace each of the matched temporal indicators with a \textbf{$\langle$mask$\rangle$} token. The upper block in Figure~\ref{fig:method} shows two examples where ``\textit{following}'' and ``\textit{after}'' are masked from the original texts.
\vspace{-.3cm}
\paragraph{Event Masks.} We build highly accurate event detection models \cite{han-etal-2019-joint, zhangTemporalGraph} to automatically label event trigger words in the 10 million passages mentioned above. Similarly, we replace these events with \textbf{$\langle$mask$\rangle$} tokens. The bottom block in Figure~\ref{fig:method} shows two examples where events ``\textit{transfer}'' and ``\textit{resumed}'' are masked from the original texts.
\vspace{-.3cm}
\subsection{Generator for Mask Predictions}
To learn effectively from the targeted samples, we train two generators with shared representations to recover temporal and event masks.
\vspace{-0.2cm}
\paragraph{Temporal Generator.} The per-sample temporal mask prediction objective is computed using cross-entropy loss,
\vspace{-.3cm}
\begin{align} \label{eq:Obj-temp}
\small
\mathcal{L}_{\mathcal{T}} = -\sum_{x^\mathcal{T}_t \in \boldsymbol{x}^\mathcal{T}} \mathcal{I}[x^\mathcal{T}_t = x_t] \log(p(x^\mathcal{T}_t|\boldsymbol{x}))
\end{align}

where $p(x^\mathcal{T}_t|\boldsymbol{x}) = Softmax\left(f_{\mathcal{T}}(h_G(\boldsymbol{x})_t)\right)$ and $x^\mathcal{T}_t \in \boldsymbol{x}^\mathcal{T}$ is a masked temporal indicator. $h_G(\boldsymbol{x})$ is $\boldsymbol{x}$'s encoded representation using a transformer and $f_{\mathcal{T}}$ is a linear layer module that maps the masked token representation into label space $\mathcal{T}$ consisting of the 40 temporal indicators.
\vspace{-0.2cm}
\paragraph{Event Generator.} The per-sample event mask prediction objective is also computed using cross-entropy loss,
\vspace{-.3cm}
\begin{align} \label{eq:Obj-temp}
\small
\mathcal{L}_{\mathcal{E}} = -\sum_{x^\mathcal{E}_t \in \boldsymbol{x}^\mathcal{E}} \mathcal{I}[x^\mathcal{E}_t = x_t] \log(p(x^\mathcal{E}_t|\boldsymbol{x}))
\vspace{-.3cm}
\end{align}
where $p(x^\mathcal{E}_t|\boldsymbol{x}) = Softmax\left(f_{\mathcal{E}}(h_G(\boldsymbol{x})_t)\right)$ and $x^\mathcal{E}_t \in \boldsymbol{x}^\mathcal{E}$ are masked events. $h_G(\boldsymbol{x})$ is the shared transformer encoder as in the temporal generator and $f_{\mathcal{E}}$ is a linear layer module that maps the masked token representation into label space $\mathcal{E}$ which is a set of all event triggers in the data.

\subsection{Discriminator for Contrastive Learning}
We incorporate a discriminator that provides additional feedback on mask predictions, which helps correct errors made by the generators.
\vspace{-0.2cm}
\paragraph{Contrastive Loss.}
For a masked token $x_t$, we design a discriminator to predict whether the recovered token by the mask prediction is original or corrupted. As shown in Figure~\ref{fig:method}, ``\textit{following}'' and ``\textit{resumed}'' are predicted correctly, so they are labeled as \textbf{original} whereas ``\textit{during}'' and ``\textit{run}'' are incorrectly predicted and labeled as \textbf{corrupted}. We train the discriminator with a contrastive loss,
\vspace{-0.3cm}
\begin{align} \label{eq:Obj-discriminator}
\small
\mathcal{L}_D = -\sum_{x_t \in M} y\log(D(x_t|\boldsymbol{x})) + (1-y)\log(1 -D(x_t|\boldsymbol{x})) \nonumber
\end{align}
where $M = \boldsymbol{x}^\mathcal{E}\cup\boldsymbol{x}^\mathcal{T}$ and $D(x_t|\boldsymbol{x}) = Sigmoid\left(f_D(h_D(\boldsymbol{x})_t)\right)$ and $y$ is a binary indicator of whether a mask prediction is correct or not. $h_D$ shares the same transformer encoder with $h_G$.
\vspace{-0.2cm}
\paragraph{Perturbed Samples.}  Our mask predictions focus on temporal and event tokens, which are easier tasks than the original mask predictions in PTLMs. This could make the contrastive loss not so powerful as training a good discriminator requires relatively balanced original and corrupted samples. To deal with this issue, for $r\%$ of the generator's output, instead of using the recovered tokens, we replace them with a token randomly sampled from either the temporal lexicon or the event vocabulary. We fix $r = 50$ to make original and corrupted samples nearly balanced.

\subsection{Joint Training}

To optimize the combining impact of all components in our model, the final training loss calculates the weighted sum of each individual loss, $\mathcal{L} = \mathcal{L}_\mathcal{T} + \alpha \mathcal{L}_\mathcal{E} + \beta\mathcal{L}_{D}$, where $\alpha$ and $\beta$ are hyper-parameters that balance different training objectives. The temporal and event masked samples are assigned a unique identifier (1 for temporal, 0 for event) so that the model knows which linear layers to feed the output of transformer into. Our overall generator-discriminator architecture resembles ELECTRA \citep{Clark2020ELECTRA}. However, our proposed method differs from this work in 1) we use targeted masking strategy as opposed to random masks; 2) both temporal and event generators and the discriminator, i.e. $h_G$ and $h_D$ share the hidden representations, but we allow task-specific final linear layers $f_\mathcal{T}$, $f_\mathcal{E}$ and $f_D$; 3) we do not train from scratch and instead continuing to train transformer parameters provided by PTLMs.

\subsection{Fine-tuning on Target Tasks}
\label{sec:finetune}
\begin{figure}[t]
\centering

    \begin{subfigure}[b]{\columnwidth}
    \includegraphics[trim=5cm 0cm 5cm 0cm, clip, angle=-90, width=0.9\columnwidth]{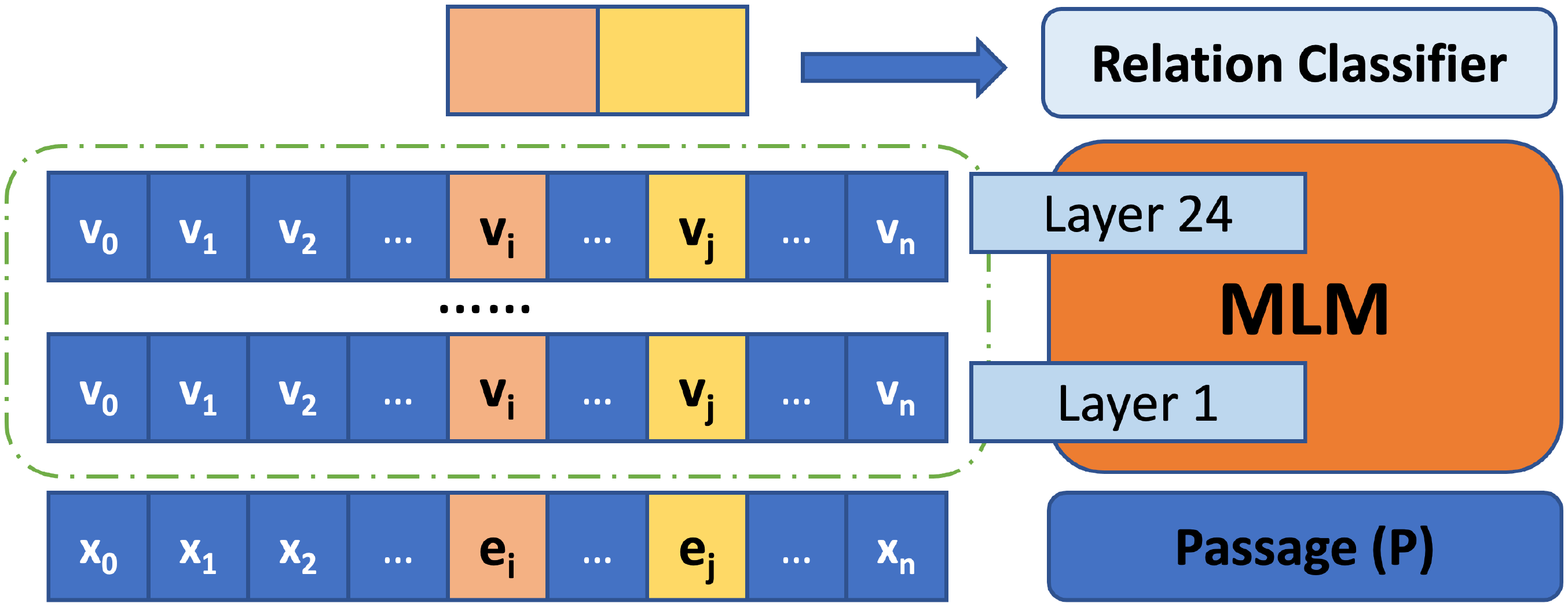}
    \vspace{-0.2cm}
    \caption{ERE}
    \label{fig:task-ere}
    \end{subfigure}
    ~
    \begin{subfigure}[b]{\columnwidth}
    \includegraphics[trim=6cm 0cm 6cm 0cm, clip, angle=-90, width=0.9\columnwidth]{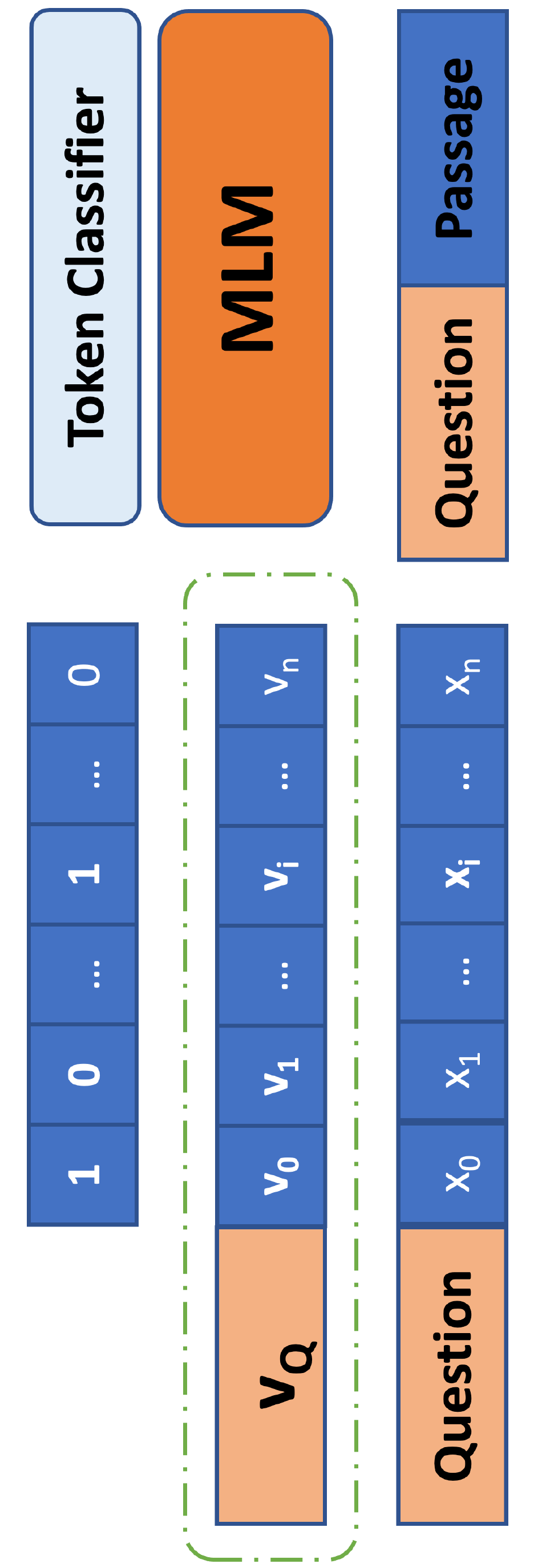}
    \vspace{-0.2cm}
    \caption{QA: {\torque}}
    \label{fig:task-torque}
    \end{subfigure}
    ~
    \begin{subfigure}[b]{\columnwidth}
    \includegraphics[trim=6cm 0cm 6cm 0cm, clip, angle=-90, width=0.9\columnwidth]{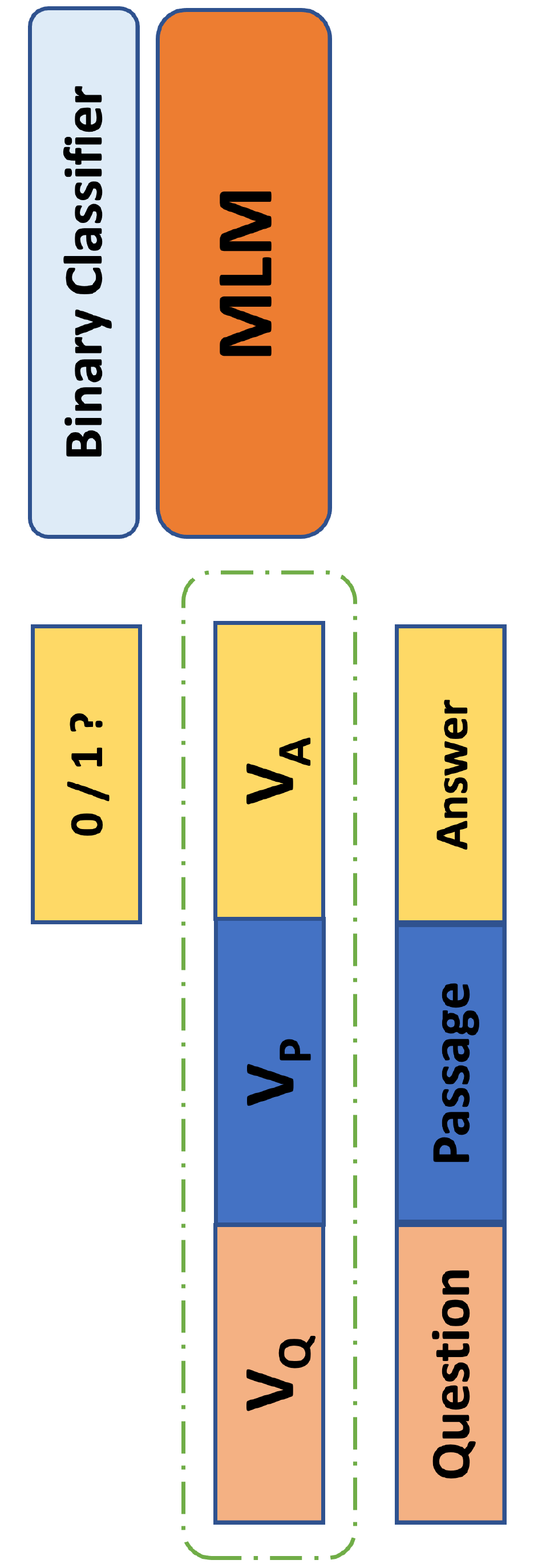}
    \vspace{-0.2cm}
    \caption{QA: {\mctaco}}
    \label{fig:task-mctaco}
    \end{subfigure}
    
\vspace{-0.2cm}
\caption{Target ERE and QA task illustrations.}
\label{fig:task-desc} 
\vspace{-0.7cm}
\end{figure}

After training with {\econet}, we fine-tune the updated MLM on the downstream tasks. ERE samples can be denoted as $\left[P, e_i, e_j, r_{i,j}\right]$, where P is the passage and $(e_i, e_j)$ is a pair of event trigger tokens in P. As Figure~\ref{fig:task-ere} shows, we feed $(P, e_i, e_j)$ into an MLM (trained with {\econet}). Following the setup of \citet{han-etal-2019-deep} and \citet{zhangTemporalGraph}, we concatenate the final event representations $v_i, v_j$ associated with $(e_i, e_j)$ to predict temporal relation $r_{i,j}$. The relation classifier is implemented by a multi-layer perceptron (MLP).

MRC/QA samples can be denoted as $\left[P, Q, A\right]$, where Q represents a question and A denotes answers. Figure~\ref{fig:task-torque} illustrates an extractive QA task where we feed the concatenated $\left[P, Q\right]$ into an MLM. Each token $x_i \in P$ has a label with 1 indicating $x_i \in A$ and 0 otherwise. The token classifier implemented by MLP predicts labels for all $x_i$. Figure~\ref{fig:task-mctaco} illustrates another QA task where A is a candidate answer for the question. We feed the concatenated $\left[P, Q, A\right]$ into an MLM and the binary classifier predicts a 0/1 label of whether A is a true statement for a given question.

\section{Experimental Setup} 
\label{sec:experiments}
In this section, we describe details of implementing {\econet}, datasets and evaluation metrics, and discuss compared methods reported in Section~\ref{sec:result}.

\subsection{Implementation Details}
\paragraph{Event Detection Model.} As mentioned briefly in Section~\ref{sec:method}, we train a highly accurate event prediction model to mask event (triggers). We experimented with two models using event annotations in {\torque} \cite{ning-etal-2020-torque} and {\tbd} \cite{ChambersTBS2014}. These two event annotations both follow previous event-centric reasoning research by using a trigger word (often a verb or an noun that most clearly describes the event’s occurrence) to represent an event \citep{uzzaman-etal-2013-semeval, glavas-etal-2014-hieve, ogorman-etal-2016-richer}. In both cases, we fine-tune {\roberta} on the \textbf{train set} and select models based on the performance on the \textbf{dev set}. The primary results shown in Table~\ref{tab:results} uses {\torque}'s annotations, but we conduct additional analysis in Section~\ref{sec:result} to show both models produce comparable results.

\paragraph{Continual Pretraining.} We randomly selected only 200K out of 10 million samples to speed up our experiments and found the results can be as good as using a lot more data. We used half of these 200K samples for temporal masked samples and the other half for the event masked samples. We ensure none of these sample passages overlap with the target \textbf{test data}. To keep the mask tokens balanced in the two training samples, we masked only 1 temporal indicator or 1 event (closest to the temporal indicator). We continued to train BERT and RoBERTa up to 250K steps with a batch size of 8. The training process takes 25 hours on a single GeForce RTX 2080 GPU with 11G memory. Note that our method requires much fewer samples and is more computation efficient than the full-scale pre-training of language models, which typically requires multiple days of training on multiple large GPUs / TPUs.

For the generator only models reported in Table~\ref{tab:results}, we excluded the contrastive loss, trained models with a batch size of 16 to fully utilize GPU memories. We leveraged the \textbf{dev set} of {\torque} to find the best hyper-parameters.

\begin{table*}[htbp!]
\centering
\scalebox{0.7}{
\setlength{\tabcolsep}{4pt}
\begin{tabular}{l|ccc|cc|c|c|c}
\toprule
 & \multicolumn{3}{c|}{\textbf{\torque}} &\multicolumn{2}{c|}{\textbf{\mctaco}} & {\textbf{\tbd}} & {\textbf{\matres}} & {\textbf{\red}}\\
\cmidrule(lr){2-4}\cmidrule(lr){5-6}\cmidrule(lr){7-7}\cmidrule(lr){8-8}\cmidrule(lr){9-9}
\textbf{Methods} & F$_1$ & EM & C & F$_1$ & EM & F$_1$ & F$_1$ &  F$_1$ \\
\midrule
TacoLM & 65.4($\pm$0.8) & 37.1($\pm$1.0) & 21.0($\pm$0.8) & 
69.3($\pm$0.6) & 40.5($\pm$0.5) & 64.8($\pm$0.7) & 70.9($\pm$0.3) & 40.3($\pm$1.7)\\
\midrule
\textbf{\bert} & 70.6($\pm1.2$) & 43.7($\pm$1.6) & 27.5($\pm$1.2) & 
70.3($\pm$0.9) & 43.2($\pm$0.6) & 62.8($\pm$1.4) & 70.5($\pm$0.9) & 39.4($\pm0.6$)\\
\textbf{+ {\econet}} & 71.4($\pm$0.7) & 44.8($\pm$0.4) & 28.5($\pm$0.5) & 
69.2($\pm$0.9) & 42.3($\pm$0.5) & 63.0($\pm$0.6) & 70.4($\pm$0.9) & 40.2($\pm0.8$) \\
\midrule
\textbf{\roberta} & 75.1($\pm$0.4) & 49.6($\pm$0.5) & 35.3($\pm$0.8) & 75.5($\pm$1.0) & 50.4($\pm$0.9) & 62.8($\pm$0.3) & 78.3($\pm$0.5) & 39.4($\pm$0.4)\\
\textbf{+ Generator} & 75.8($\pm$0.4) & 51.2($\pm$1.1) & 35.8($\pm$0.9) & 75.1($\pm$1.4) & 50.2($\pm$1.2) & 65.2($\pm$0.6) & 77.0($\pm$0.9) & 41.0($\pm$0.6) \\
\textbf{+ {\econet}} & 76.1($\pm$0.2) & 51.6($\pm$0.4) & 36.8($\pm$0.2) & 76.3($\pm$0.3) & 52.8($\pm$1.9) & 64.8($\pm$1.4) & 78.8($\pm$0.6) & 42.8($\pm$0.7) \\
\midrule
{\econet} (best) & \textbf{76.3} & \textbf{52.0} & \textbf{37.0} & 76.8 & 54.7 & \textbf{66.8} & 79.3 & \textbf{43.8} \\
{Current SOTA} & 75.2$^*$ & 51.1 & 34.5 & \textbf{79.5}$^\dagger$ & \textbf{56.5} & 66.7$^{\dagger\dagger}$& \textbf{80.3}$^\ddagger$  & 34.0$^{\ddagger\ddagger}$ \\
\bottomrule
\end{tabular}
}
\vspace{-0.2cm}
\caption{Overall experimental results. Refer to Section~\ref{sec:naming} for naming conventions. The SOTA performances for \torque$^*$  are provided by \citet{ning-etal-2020-torque} and the numbers are average over 3 random seeds. The SOTA performances for \mctaco$^\dagger$ are provided by \citet{pereira-etal-2020-adversarial}; \tbd$^{\dagger\dagger}$ and \matres$^\ddagger$ by \citet{zhangTemporalGraph} and \red$^{\ddagger\ddagger}$ by \citet{contextualTemporal-han}. $^\dagger$, $^{\dagger\dagger}$, $^\ddagger$ and $^{\ddagger\ddagger}$ only report the best single model results, and to make fair comparisons with these baselines, we report both average and best single model performances. TacoLM baseline uses the provided and recommended checkpoint for extrinsic evaluations.} 
\label{tab:results}
\vspace{-0.6cm}
\end{table*}
\vspace{-0.2cm}

\paragraph{Fine-tuning.} Dev set performances were used for early-stop and average \textbf{dev performances} over three randoms seeds were used to pick the best hyper-parameters. Note that \textbf{test set} for the target tasks were never observed in any of the training process and their performances are reported in Table~\ref{tab:results}. All hyper-parameter search ranges can be found in Appendix~\ref{sec:reproduce}.

\vspace{-0.2cm}
\subsection{Datasets} 
We evaluate our approach on five datasets concerning temporal ERE and MRC/QA. We briefly describe these data below and list detailed statistics in Appendix~\ref{sec:data-summary}.

\paragraph{ERE Datasets.} {\tbd} \citep{ChambersTBS2014}, {\matres} \citep{NingWuRo18} and {\red} \citep{ogorman-etal-2016-richer} are all ERE datasets. Their samples follow the input format described in Section~\ref{sec:finetune} where a pair of event (triggers) together with their context are provided. The task is to predict pairwise event temporal relations. The differences are how temporal relation labels are defined. Both {\tbd} and {\matres} leverage a {\temprel{vague}} label to capture relations that are hard to determine even by humans, which results in denser annotations than {\red}. {\red} contains the most fine-grained temporal relations and thus the lowest sample/relation ratio. {\matres} only considers start time of events to determine their temporal order, whereas {\tbd} and {\red} consider start and end time, resulting in lower inter-annotator agreement.

\paragraph{{\torque}} \citep{ning-etal-2020-torque} is an MRC/QA dataset where annotators first identify event triggers in given passages and then ask questions regarding event temporal relations (ordering). Correct answers are event trigger words in passages. {\torque} can be considered as reformulating temporal ERE tasks as an MRC/QA task. Therefore, both ERE datasets and {\torque} are highly correlated with our continual pre-training objectives where targeted masks of both events and temporal relation indicators are incorporated.

\paragraph{\mctaco} \citep{zhou-etal-2019-going} is another MRC/QA dataset, but it differs from {\torque} in 1) events are not explicitly identified; 2) answers are statements with true or false labels; 3) questions contain broader temporal commonsense regarding not only temporal ordering, but also event frequency, during and typical time that may not be directly helpful for reasoning temporal relations. For example, knowing how often a pair of events happen doesn't help us figure out which event happens earlier. Since our continual pre-training focuses on temporal relations, {\mctaco} could the least compatible dataset in our experiments.

\vspace{-0.2cm}
\subsection{Evaluation Metrics}
Three metrics are used to evaluate the fine-tuning performances.
\vspace{-0.2cm}
\paragraph{$\mathbf{F_1}$:} for {\torque} and {\mctaco}, we follow the data papers \citep{ning-etal-2020-torque} and \citep{zhou-etal-2019-going} to report macro average of each question's $F_1$ score. For {\tbd}, {\matres} and {\red}, we report standard micro-average $F_1$ scores to be consistent with the baselines.
\vspace{-0.3cm}
\paragraph{Exact-match (EM):} for both MRC datasets, EM = 1 if answer predictions match perfectly with gold annotations; otherwise, EM = 0.
\vspace{-0.3cm}
\paragraph{EM-consistency (C):} in {\torque}, some questions can be clustered into the same group due to the data collection process. This metric reports the average EM score for a group as opposed to a question in the original EM metrics.
\vspace{-0.1cm}
\subsection{Compared Methods} 
\label{sec:naming}
We compare several pre-training methods with {\econet}: 1) \textbf{{\roberta}} is the original PTLM and we fine-tune it directly on target tasks; 2) \textbf{{\roberta} + {\econet}} is our proposed continual pre-training method; 3) \textbf{{\roberta} + Generator} only uses the generator component in continual pre-training; 4) \textbf{{\roberta} + random mask} keeps the original PTLMs' objectives and replaces the targeted masks in {\econet} with randomly masked tokens. The methods' names for continual pre-training {\bert} can be derived by replacing {\roberta} with {\bert}. 

We also fine-tune pre-trained TacoLM on target datasets. The current SOTA systems we compare with are provided by \citet{ning-etal-2020-torque}, \citet{pereira-etal-2020-adversarial}, \citet{zhangTemporalGraph} and \citet{contextualTemporal-han}. More details are presented in Section~\ref{sec:result-sota}.
\vspace{-0.6cm}

\section{Results and Analysis}
\label{sec:result}
\vspace{-0.2cm}
As shown in Table~\ref{tab:results}, we report two baselines. The first one, TacoLM is a related work that focuses on event duration, frequency and typical time. The second one is the current SOTA results reported to the best of the authors' knowledge. We also report our own implementations of fine-tuning {\bert} and {\roberta} to compare fairly with \econet. Unless pointing out specifically, all gains mentioned in the following sections are in the unit of \textbf{absolute percentage}. 
\vspace{-0.2cm}
\subsection{Comparisons with Existing Systems}
\label{sec:result-sota}
\paragraph{\torque.} The current SOTA system reported in \citet{ning-etal-2020-torque} fine-tunes {\roberta} and our own fine-tuned {\roberta} achieves on-par $\mathbf{F_1}$, EM and \textbf{C} scores. The gains of {\roberta} + {\econet} against the current SOTA performances are 0.9\%, 0.5\% and 2.3\% per $F_1$, EM and \textbf{C} metrics.
\vspace{-0.2cm}
\paragraph{\mctaco.} The current SOTA system ALICE \citep{pereira-etal-2020-adversarial} also uses {\roberta} as the text encoder, but leverages adversarial attacks on input samples. ALICE achieves 79.5\% and 56.5\% per $F_1$ and EM metrics on the test set for the best single model, and the best performances for {\roberta} + {\econet} are 76.8\% and 54.7\% per $F_1$ and EM scores, which do not outperform ALICE. This gap can be caused by the fact that the majority of samples in {\mctaco} reason about event frequency, duration and time, which are not directly related to event temporal relations.
\vspace{-0.3cm}
\paragraph{{\tbd} + \matres.} The most recent SOTA system reported in \citet{zhangTemporalGraph} uses both {\bert} and {\roberta} as text encoders, but leverages syntactic parsers to build large graphical attention networks on top of PTLMs. {\roberta} + {\econet}'s fine-tuning performances are essentially on-par with this work without additional parameters. For {\tbd}, our best model outperforms \citet{zhangTemporalGraph} by 0.1\% while for {\matres}, our best model underperforms by 1.0\% per $F_1$ scores.
\vspace{-0.3cm}
\paragraph{\red.} The current SOTA system reported in \citet{contextualTemporal-han} uses BERT$_{BASE}$ as word representations (no finetuning) and BiLSTM as feature extractor. The single best model achieves 34.0\% $F_1$ score and {\roberta} + {\econet} is 9.8\% higher than the baseline.
\vspace{-0.2cm}
\subsection{The Impact of {\econet}} 
\vspace{-0.2cm}
\paragraph{Overall Impact.} {\econet} in general works better than the original {\roberta} across 5 different datasets, and the improvements are more salient in {\torque} with 1.0\%, 2.0\% and 1.5\% gains per $F_1$, EM and \textbf{C} scores, in {\mctaco} with 2.4\% lift over the EM score, and in {\tbd} and {\red} with 2.0\% and 3.4\% improvements respectively over $F_1$ scores. We observe that the improvements of {\econet} over {\bert} is smaller and sometimes hurts the fine-tuning performances. We speculate this could be related to the property that BERT is less capable of handling temporal reasoning tasks, but we leave more rigorous investigations to future research.
\vspace{-0.2cm}
\paragraph{Impact of Contrastive Loss.} Comparing the average performances of continual pre-training with generator only and with {\econet} (generator + discriminator), we observe that generator alone can improve performances of {\roberta} in 3 out of 5 datasets. However, except for {\tbd}, {\econet} is able to improve fine-tuning performances further, which shows the effectiveness of using the contrastive loss.
\vspace{-0.2cm}
\paragraph{Significance Tests.} As current SOTA models are either not publicly available or under-perform {\roberta}, we resort to testing the statistical significance of the best single model between {\econet} and {\roberta}. Table~\ref{tab:significance} in the appendix lists all improvements' p-values per McNemar's test \citep{McNemar-1947}. {\matres} appears to be the only one that is not statistically significant.
\vspace{-0.7cm}
\subsection{Impact of Event Models}
\vspace{-0.1cm}
Event trigger definitions have been consistent in previous event temporal datasets \cite{ogorman-etal-2016-richer, ChambersTBS2014, ning-etal-2020-torque}. Trigger detection models built on {\torque} and {\tbd} both achieve $>92\%$ $F_1$ scores and $>95\%$ precision scores. For the 100K pre-training data selected for event masks, we found an 84.5\% overlap of triggers identified by both models. We further apply {\econet} trained on both event mask data to the target tasks and achieve comparable performances shown in Table~\ref{tab:event-model-impact} of the appendix. These results suggest that the impact of different event annotations is minimal and triggers detected in either model can generalize to different tasks.

\begin{table}[h!]
\centering
\setlength{\tabcolsep}{3.0pt}
\scalebox{0.65}{
\begin{tabular}{lccc|c|c}
\toprule
& \multicolumn{3}{c|}{\textbf{\torque}} & {\textbf{TB-D}} & {\textbf{\red}}\\
\cmidrule(lr){2-4}\cmidrule(lr){5-5}\cmidrule(lr){6-6}
\textbf{Methods} & F$_1$ & EM & C & F$_1$ & F$_1$ \\
\midrule
\textbf{{\roberta}} & 75.1 & 49.6 & \underline{35.3} & 62.8 & 39.4\\
\textbf{+ random mask} & 74.9& 49.5& \underline{35.1}& \underline{58.7} & \underline{38.3}\\
\textbf{+ {\econet}} &  \textbf{76.1} & \textbf{51.6} & \textbf{36.8} & \underline{\textbf{64.8}} & \textbf{42.8} \\
\bottomrule
\end{tabular}}
\vspace{-0.2cm}
\caption{Fine-tuning performances with different pre-training methods. All numbers are average over 3 random seeds. Std. Dev. $\ge 1\%$ is underlined.}
\label{tab:abl-result}
\vspace{-0.7cm}
\end{table}

\subsection{Additional Ablation Studies} 
\vspace{-0.1cm}
To better understand our proposed model, we experiment with additional continual training methods and compare their fine-tuning performances. 
\vspace{-0.3cm}
\paragraph{Random Masks.} As most target datasets we use are in the news domain, to study the impact of potential domain-adaption, we continue to train PTLMs with the original objective on the same data using random masks. To compare fairly with the generator and {\econet}, we only mask 1 token per training sample. The search range of hyper-parameters is the same as in Section~\ref{sec:experiments}. As Table~\ref{tab:abl-result} and \ref{tab:abl-result-bert} (appendix) show, continual pre-training with random masks, in general, does not improve and sometimes hurt fine-tuning performances compared with fine-tuning with original PTLMs. We hypothesize that this is caused by masking a smaller fraction (1 out of $\approx$50 average) tokens than the original 15\%. {\roberta} + {\econet} achieves the best fine-tuning results across the board.

\begin{table}[h!]
\centering
\vspace{-0.2cm}
\scalebox{0.65}{
\setlength{\tabcolsep}{3.7pt}
\begin{tabular}{l|ccc|ccc}
\toprule
&\multicolumn{3}{c|}{Full Train Data} & \multicolumn{3}{c}{10\% Train Data} \\
\midrule
& RoBERTa & $\Delta$ & $\Delta \%$ & RoBERTa & $\Delta$ & $\Delta \%$ \\ 
\midrule
 {\textbf{\torque}} & 75.1 & +1.0 & +1.3\% & 59.7 & \textbf{+7.2} & \textbf{+12.1\%} \\ 
 {\textbf{\mctaco}} & 75.5 & +0.8 & +1.1\% & 44.0 & \textbf{+5.6} & \textbf{+12.7\%} \\
 {\textbf{\tbd}} & 62.8 & +2.0 & +3.2\% & 48.8 & \textbf{+2.8} & \textbf{+5.7\%} \\
 {\textbf{\matres}} & 78.3 & +0.5 & +1.3\% & 71.0 & \textbf{+2.4} & \textbf{+3.4\%} \\
 {\textbf{\red}} & 39.4 & \textbf{+2.4} & +6.0\% & 27.2 & +1.8 & \textbf{+6.6\%} \\
\bottomrule
\end{tabular}
}
\vspace{-0.2cm}
\caption{{\roberta} + {\econet}'s improvements over {\roberta} using full train data v.s. 10\% of train data. $\Delta$ indicates absolute points improvements while $\Delta \%$ indicates relative gains per $F_1$ scores.} 
\label{tab:low-resource}
\vspace{-0.8cm}
\end{table}

\subsection{Fine-tuning under Low-resource Settings}
\vspace{-0.1cm}
In Table~\ref{tab:low-resource}, we compare the improvements of fine-tuning {\roberta} + {\econet} over {\roberta} using full and 10\% of the training data. Measured by both absolute and relative percentage gains, the majority of the improvements are much more significant under low-resource settings. This suggests that the transfer of event temporal knowledge is more salient when data is scarce. We further show fine-tuning performance comparisons using different ratios of the training data in Figure~\ref{fig:torque-ratio}-\ref{fig:red-ratio} in the appendix. The results demonstrate that {\econet} can outperform {\roberta} consistently when fine-tuning {\torque} and {\red}.
\vspace{-0.5cm}
\subsection{Attention Scores on Temporal Indicators}
\vspace{-0.2cm}
In this section, we attempt to show explicitly how {\econet} enhances MLMs' attentions on temporal indicators for downstream tasks. As mentioned in Sec.~\ref{sec:finetune}, for a particular ERE task (e.g. {\tbd}), we need to predict the temporal relations between a pair of event triggers $e_i, e_j \in P_{i,j}$ with associated vector representations $v_i^{l, h}, v_j^{l, h}, l \in L, h \in H$ in an MLM. $L$ and $H$ are the number of layers and attention heads respectively. We further use $T_m \in T$ to denote a temporal indicator category listed in Table~\ref{tab:temporal-indicators}, and $t_{m, n} \in T_m$ denote a particular temporal indicator. If we let $attn(v_i^{l, h}, v_x^{l, h})$ represents the attention score between an event vector and any other hidden vectors, we can aggregate the per-layer attention score between $e_i$ and $t_{m, n}$ as,
$
a^l_{i, {t_{m, n}}} = \frac{1}{H} \sum_h^H attn(v_i^{l, h}, v_{t_{m, n}}^{l, h})
$.
Similarly, we can compute $a^l_{j, {t_{m, n}}}$. The final per-layer attention score for $(e_i, e_j)$ is
$
a^l_{t_{m, n}} = \frac{1}{2} \left( a^l_{i, {t_{m, n}}} + a^l_{j, {t_{m, n}}} \right)
$.
To compute the attention score for the $T_m$ category, we take the average of $\{a^l_{t_{m, n}} \mid \forall t_{m, n} \in T_m \mbox{ and } \forall t_{m, n} \in P_{i, j}\}$. Note we assume a temporal indicator is a single token to simplify notations above; for multiple-token indicators, we take the average of $attn(v_i^{l, h}, v_{x \in t_{m,n}}^{l, h})$.
\begin{figure}[t]
    \centering
\includegraphics[width=0.8\columnwidth]{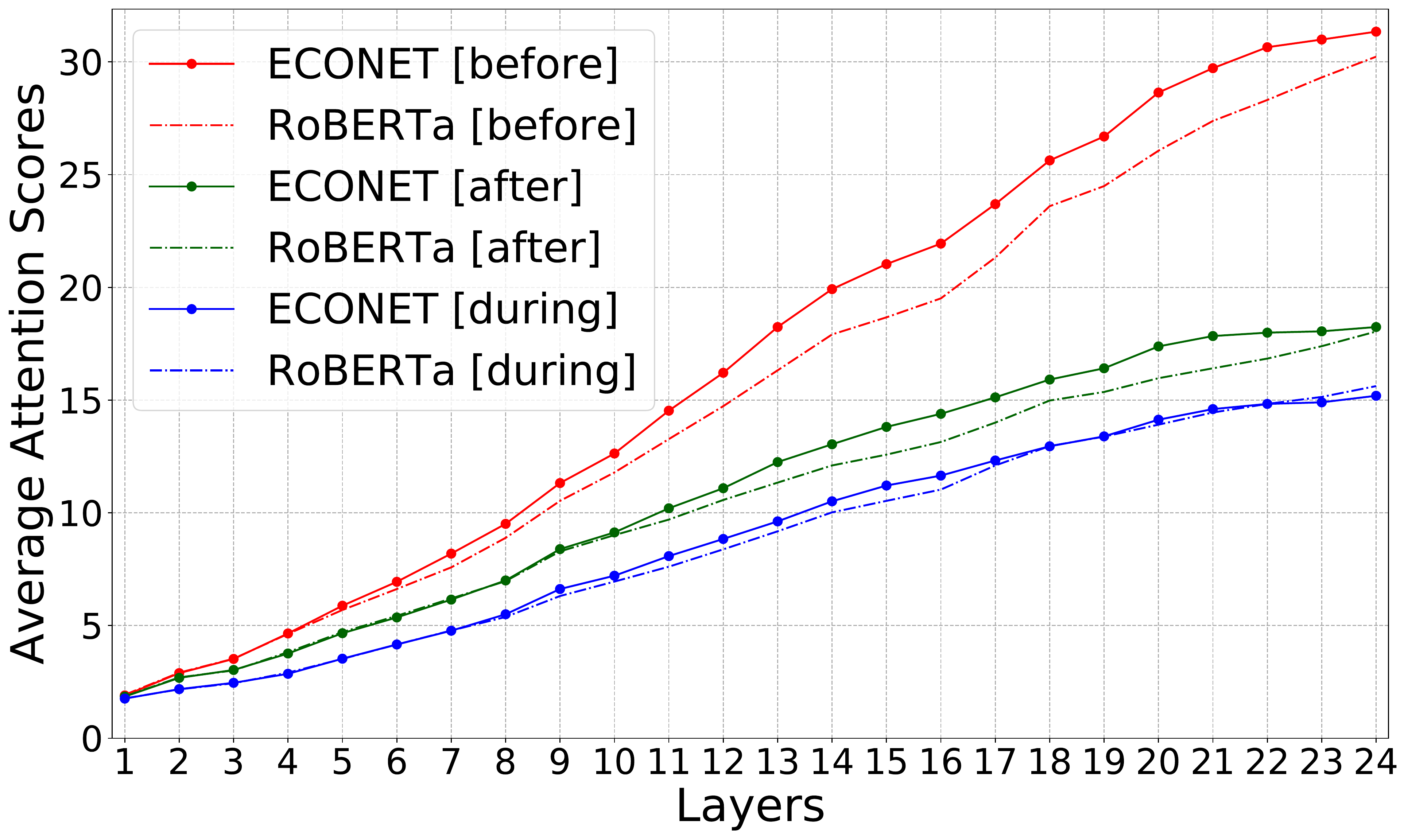}
\vspace{-0.4cm}
\caption{Cumulative attention score comparisons between {\roberta} and {\econet} on {\tbd} test data. All numbers are multiplied by 100 and averaged over 3 random seeds for illustration clarity.}
\label{fig:tbd-attn}
\vspace{-0.7cm}
\end{figure}

Figure~\ref{fig:tbd-attn} shows the cumulative attention scores for temporal indicator categories, \textbf{[before]}, \textbf{[after]} and \textbf{[during]} in ascending order of model layers. We observe that the attention scores for {\roberta} and {\econet} align well on the bottom layers, but {\econet} outweighs {\roberta} in middle to top layers. Previous research report that upper layers of pre-trained language models focus more on complex semantics as opposed to shallow surface forms or syntax on the lower layers \citep{tenney-etal-2019-bert, jawahar-etal-2019-bert}. Thus, our findings here show another piece of evidence that targeted masking is effective at capturing temporal indicators, which could facilitate semantics tasks including temporal reasoning.

\vspace{-0.1cm}
\subsection{Temporal Knowledge Injection}
\vspace{-0.2cm}
We hypothesize in the introduction that vanilla PTLMs lack special attention to temporal indicators and events, and our proposed method addresses this issue by a particular design of mask prediction strategy and a discriminator that is able to distinguish reasonable events and temporal indicators from noises. In this section, we show more details of how such a mechanism works. 

\begin{figure}[t]
\centering

    \begin{subfigure}[b]{\columnwidth}
    \includegraphics[trim=0cm 0cm 0cm 0cm, clip, width=0.75\columnwidth]{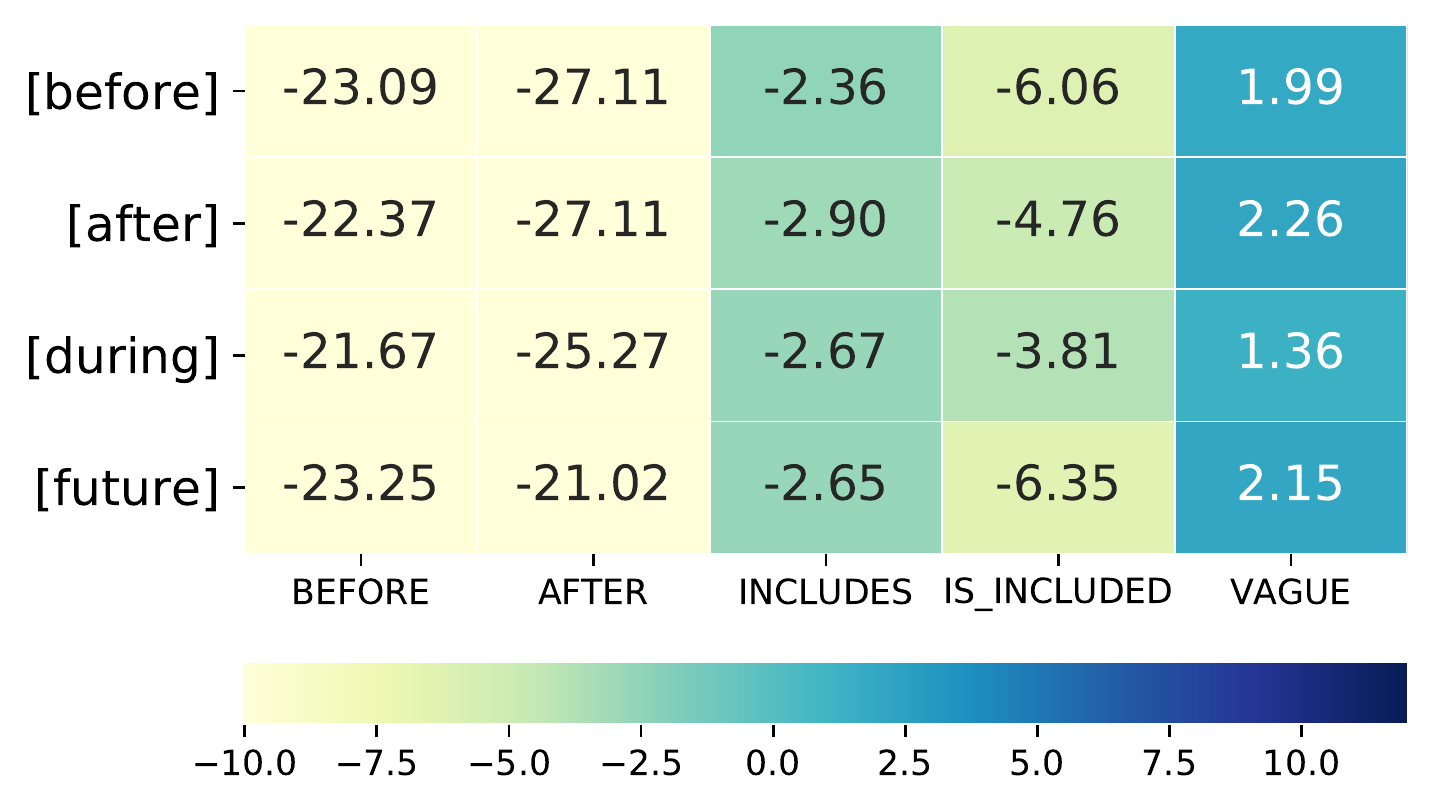}
    \vspace{-0.2cm}
    \caption{Random Mask - {\roberta}}
    \label{fig:random-roberta}
    \end{subfigure}
    
    \begin{subfigure}[b]{\columnwidth}
    \includegraphics[trim=0cm 0cm 0cm 0cm, clip, width=0.75\columnwidth]{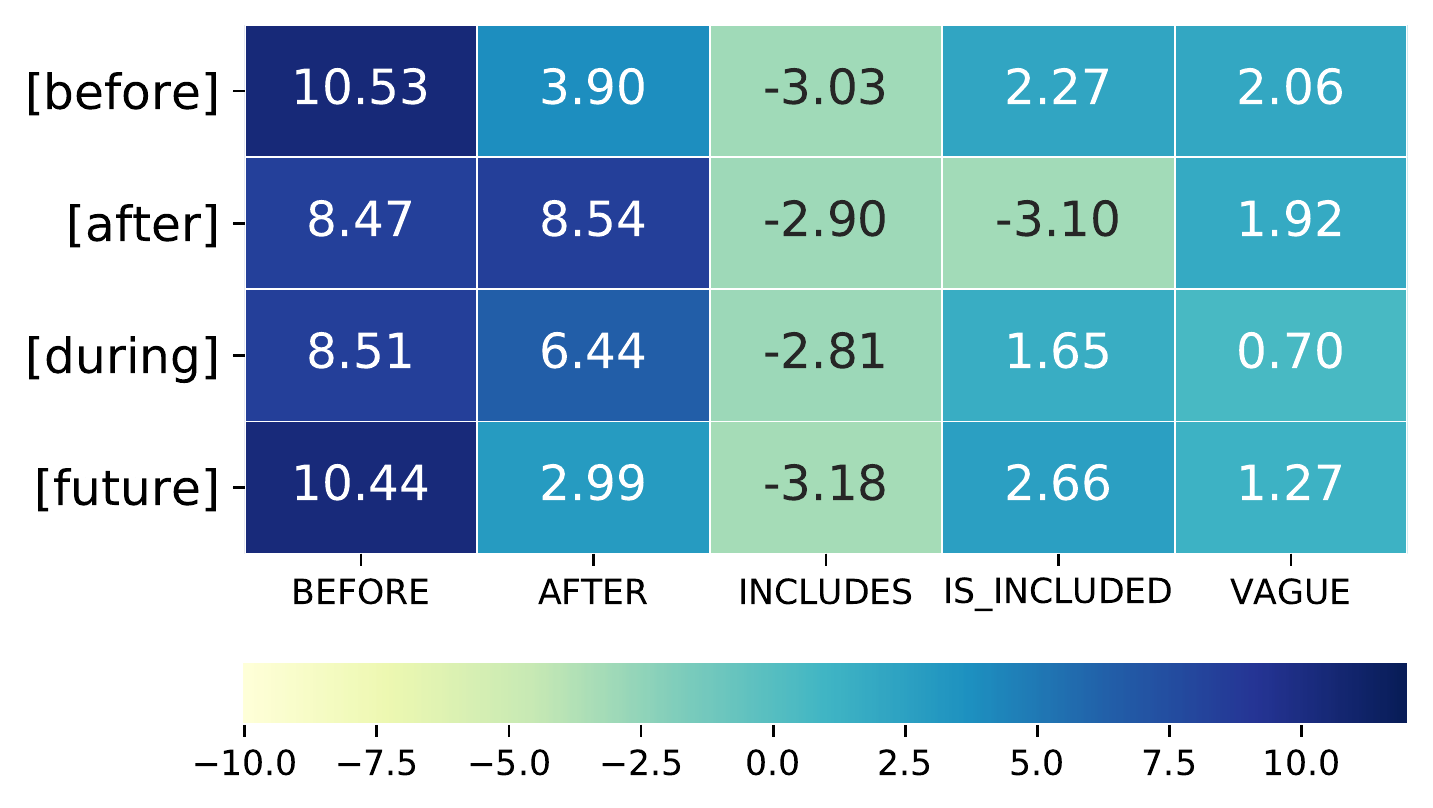}
    \vspace{-0.2cm}
    \caption{{\econet} - {\roberta}}
    \label{fig:deer-roberta}
    \end{subfigure}
    
\vspace{-0.4cm}
\caption{Performance (F$_1$ score) differences by temporal indicator categories and label classes in {\tbd}. Fine-tuning on 10\% {\tbd} training data.}
\label{fig:lifts} 
\vspace{-0.7cm}
\end{figure}

The heat maps in Figure~\ref{fig:lifts} calculate the fine-tuning performance differences between 1) {\roberta} and continual pre-training with random masks (Figure~\ref{fig:random-roberta}); and 2) between {\roberta} and {\econet}  (Figure~\ref{fig:deer-roberta}). Each cell shows the difference for each label class in {\tbd} conditional on samples' input passage containing a temporal indicator in the categories specified in Table~\ref{tab:temporal-indicators}. Categories with less than 50 sample matches are excluded from the analysis.

In Figure~\ref{fig:random-roberta}, the only gains come from \temprel{vague}, which is an undetermined class in {\tbd} to handle unclear pairwise event relations. This shows that continual pre-training with random masks works no better than original PTLMs to leverage existing temporal indicators in the input passage to distinguish positive temporal relations from unclear ones. On the other hand, in Figure~\ref{fig:deer-roberta}, having temporal indicators in general benefits much more for \temprel{before}, \temprel{after}, \temprel{is\_included} labels. The only exception is \temprel{includes}, but it is a small class with only 4\% of the data.

More interestingly, notice the diagonal cells, i.e. ([before], \temprel{before}), ([after], \temprel{after}) and ([during], \temprel{INCLUDES}) have the largest values in the respective columns. These results are intuitive as temporal indicators should be most beneficial for temporal relations associated with their categories. Combining these two sets of results, we provide additional evidence that {\econet} helps PTLMs better capture temporal indicators and thus results in stronger fine-tuning performances.

Our final analysis attempts to show why discriminator helps. We feed 1K unused masked samples into the generator of the best {\econet} in Table~\ref{tab:results} to predict either the masked temporal indicators or masked events. We then examine the accuracy of the discriminator for correctly and incorrectly predicted masked tokens. As shown in Table~\ref{tab:discriminator-res} of the appendix, the discriminator aligns well with the event generator's predictions. For the temporal generator, the discriminator disagrees substantially (82.2\%) with the ``incorrect'' predictions, i.e. the generator predicts a supposedly wrong indicator, but the discriminator thinks it looks original.

To understand why, we randomly selected 50 disagreed samples and found that 12 of these ``incorrect'' predictions fall into the same temporal indicator group of the original ones and 8 of them belong to the related groups in Table~\ref{tab:temporal-indicators}. More details and examples can be found in Table~\ref{tab:discriminator-ex} in the appendix. This suggests that despite being nearly perfect replacements of the original masked indicators, these 40\% samples are penalized as wrong predictions when training the generator. The discriminator, by disagreeing with the generator, provides opposing feedback that trains the overall model to better capture indicators with similar temporal signals.
\section{Related Work}
\label{sec:related}
\vspace{-0.2cm}
\paragraph{Language Model Pretraining.}
Since the breakthrough of BERT \citep{BERT2018}, PTLMs have become SOTA models for a variety of NLP applications. There have also been several modifications/improvements built on the original BERT model. RoBERTa \citep{ROBERTA-19} removes the next sentence prediction in BERT and trains with longer text inputs and more steps. ELECTRA \citep{Clark2020ELECTRA} proposes a generator-discriminator architecture, and addresses the sample-inefficiency issue in previous PTLMs.

Recent research explored methods to continue to train PTLMs so that they can adapt better to downstream tasks. For example, TANDA \citep{garg2019tanda} adopts an intermediate training on modified Natural Questions dataset \citep{47761} so that it performs better for the Answer Sentence Selection task. \citet{zhou2020pretraining} proposed continual training objectives that require a model to distinguish natural sentences from those with concepts randomly shuffled or generated by models, which enables language models to capture large-scale commonsense knowledge.
\vspace{-0.2cm}
\paragraph{Event Temporal Reasoning.}
There has been a surge of attention to event temporal reasoning research recently. Some noticeable datasets include ERE samples: {\tbd} \citep{ChambersTBS2014}, {\matres} \citep{NingWuRo18} and {\red} \citep{ogorman-etal-2016-richer}. Previous SOTA systems on these data leveraged PTLMs and structured learning \citep{han-etal-2019-joint, wang-etal-2020-joint, clinicalTemporal-zhou,han-etal-2020-domain} and have substantially improved model performances, though none of them tackled the issue of lacking event temporal knowledge in PTLMs. {\torque} \citep{ning-etal-2020-torque} and {\mctaco} \citep{zhou-etal-2019-going} are recent MRC datasets that attempt to reason about event temporal relations using natural language rather than ERE formalism. 

\citet{ZNKR20} and \citet{Zhao2020Temporal} are two recent works that attempt to incorporate event temporal knowledge in PTLMs. The formal one focuses on injecting temporal commonsense with targeted event time, frequency and duration masks while the latter one leverages distantly labeled pairwise event temporal relations, masks before/after indicators, and focuses on ERE application only. Our work differs from them by designing a targeted masking strategy for event triggers and comprehensive temporal indicators, proposing a continual training method with mask prediction and contrastive loss, and applying our framework on a broader range of event temporal reasoning tasks. 
\vspace{-0.2cm}
\section{Conclusion and Future Work}
\label{sec:conclude}
\vspace{-0.1cm}
In summary, we propose a continual training framework with targeted mask prediction and contrastive loss to enable PTLMs to capture event temporal knowledge. Extensive experimental results show that both the generator and discriminator components can be helpful to improve fine-tuning performances over 5 commonly used data in event temporal reasoning. The improvements of our methods are much more pronounced in low-resource settings, which points out a promising direction for few-shot learning in this research area.
\section*{Acknowledgments}
This work is supported by the Intelligence Advanced Research Projects Activity (IARPA), via Contract No. 2019-19051600007 and DARPA under agreement
number FA8750-19-2-0500.

\bibliographystyle{acl_natbib}
\bibliography{anthology, emnlp2021}
\clearpage
\appendix

\section{Data Summary}
\label{sec:data-summary}
Table~\ref{tab:data} describes basic statistics for target datasets used in this work. The numbers of train/dev/test samples for {\torque} and {\mctaco} are question based. There is no training set provided in {\mctaco}. So we train on the dev set and report the evaluation results on the test set following \citet{pereira-etal-2020-adversarial}. The numbers of train/dev/test samples for {\tbd}, {\matres} and {\red} refer to (event, event, relation) triplets. The standard dev set is not provide by {\matres} and {\red}, so we follow the split used in \citet{zhangTemporalGraph} and \citet{contextualTemporal-han}.

\begin{table}[htbp!]
\centering
\small
\begin{tabular}{l|ccccc}
\toprule
Data & \#Train & \#Dev & \#Test & \#Label\\
\cmidrule(lr){1-1}\cmidrule(lr){2-2}\cmidrule(lr){3-3}\cmidrule(lr){4-4}\cmidrule(lr){5-5}
{\torque} & 24,523 & 1,483 & 4,668 & 2\\
{\mctaco} & - & 3,783 & 9,442 & 2 \\
{\tbd} & 4,032 & 629 & 1,427 & 6\\
{\matres} & 5,412 & 920 & 827 & 4\\
{\red} & 3,336 & 400 & 473 & 11\\
\bottomrule
\end{tabular}
\caption{The numbers of samples for {\torque} refers to number of questions; the numbers for {\mctaco} are valid question and answer pairs; and the numbers of samples for {\tbd}, {\matres} and {\red} are all (event, event, relation) triplets.}
\label{tab:data}
\end{table}

Downloading link for the (processed) continual pretraining data is provided in the README file of the code package.

\section{Event Detection Model}
As mentioned briefly in Secion~\ref{sec:method}, we train an event prediction model using event annotations in {\torque}. We finetune {\roberta} on the training set and select models based on the performance on the dev set. The best model achieves $>92\%$ event prediction $F_1$ score with $>95\%$ precision score after just 1 epoch of training, which indicates that this is a highly accurate model.

\section{Reproduction Checklist}
\label{sec:reproduce}
\paragraph{Number of parameters.} We continue to train {\bert} and {\roberta} and so the number of parameters are the same as the original PTLMs, which is 336M.

\paragraph{Hyper-parameter Search} Due to computation constraints, we had to limit the search range of hyper-parameters for {\econet}. For learning rates, we tried $(1e^{-6}, 2e^{-6})$; for weights on the contrastive loss ($\beta$), we tried $(1.0, 2.0)$. 

\paragraph{Best Hyper-parameters.}
In Table~\ref{tab:best-hyper} and Table~\ref{tab:best-hyper-bert}, we provide hyper-parameters for our best performing language model using {\roberta} + {\econet} and {\bert} + {\econet} and best hyper-parameters for fine-tuning them on downstream tasks. For fine-tuning on the target datasets. We conducted grid search for learning rates in the range of $(5e^{-6}, 1e^{-5})$ and for batch size in the range of $(2, 4, 6, 12)$. We fine-tuned all models for 10 epochs with three random seeds $(5, 7, 23)$.

\begin{table}[htbp!]
\small
\centering
\begin{tabular}{lccc}
\toprule
Method & learning rate & batch size & $\beta$ \\
\cmidrule(lr){1-1}\cmidrule(lr){2-2}\cmidrule(lr){3-3}\cmidrule(lr){4-4}
{\econet} & ${1e^{-6}}$ & 8 & 1.0 \\
{\torque} & ${1e^{-5}}$ & 12 & - \\
{\mctaco}  & ${5e^{-6}}$ & 4 & - \\
{\tbd} &  ${5e^{-6}}$ & 4 & - \\
{\matres}  & ${5e^{-6}}$ & 2 & - \\
{\red}  & ${5e^{-6}}$ & 2 &  -\\
\bottomrule
\end{tabular}
\caption{Hyper-parameters of our best performing LM with {\roberta} + {\econet} as well as best hyper-parameters for fine-tuning on downstream tasks.}
\label{tab:best-hyper}
\end{table}

\begin{table}[htbp!]
\small
\centering
\begin{tabular}{lccc}
\toprule
Method & learning rate & batch size & $\beta$ \\
\cmidrule(lr){1-1}\cmidrule(lr){2-2}\cmidrule(lr){3-3}\cmidrule(lr){4-4}
{\econet} & ${2e^{-6}}$ & 8 & 1.0 \\
{\torque} & ${1e^{-5}}$ & 12 & - \\
{\mctaco}  & ${1e^{-5}}$ & 2 & - \\
{\tbd} &  ${1e^{-5}}$ & 2 & - \\
{\matres}  & ${5e^{-6}}$ & 4 & - \\
{\red}  & ${1e^{-5}}$ & 6 &  -\\
\bottomrule
\end{tabular}
\caption{Hyper-parameters of our best performing LM with {\bert} + {\econet} as well as best hyper-parameters for fine-tuning on downstream tasks.}
\label{tab:best-hyper-bert}
\end{table}

\paragraph{Dev Set Performances}
We show average dev set performances in Table~\ref{tab:results-dev} corresponding to our main results in Table~\ref{tab:results}.

\section{Significance Tests.}
We leverage McNemar's tests \citep{McNemar-1947} to show {\econet}'s improvements against {\roberta}. McNemar's tests compute statistics by aggregating all samples' prediction correctness. For ERE tasks, this value is simply classification correctness; for QA tasks ({\torque} and {\mctaco}), we use \textbf{EM} per question-answer pairs.

\begin{table}[htbp!]
\centering
\begin{tabular}{l|c}
\toprule
Datasets & p-values \\
\midrule
{\torque} &  0.002$^{**}$\\
{\mctaco}  & 0.007$^{**}$\\
{\tbd} &  0.004$^{**}$ \\
{\matres}  & 0.292  \\
{\red}  & 0.059$^{*}$ \\
\bottomrule
\end{tabular}
\caption{McNemar's tests for improvement significance between best single models of {\roberta} and {\econet} on the test data. Tests with p-values < 0.05 ($^{**}$) indicate strong statistical significance; tests with p-values < 0.1 ($^{*}$) indicate weak statistical significance.}
\label{tab:significance}
\end{table}

\section{Fine-tuning under Low-resource Settings}
\label{sec:low-res}
Table~\ref{tab:low-resource} shows improvements of {\econet} over {\roberta} are much more salient under low-resource setting.

\begin{figure}[h!]
\centering
    \begin{subfigure}[b]{\columnwidth}
    \includegraphics[trim=0cm 0cm 0cm 0cm, clip, width=0.9\columnwidth]{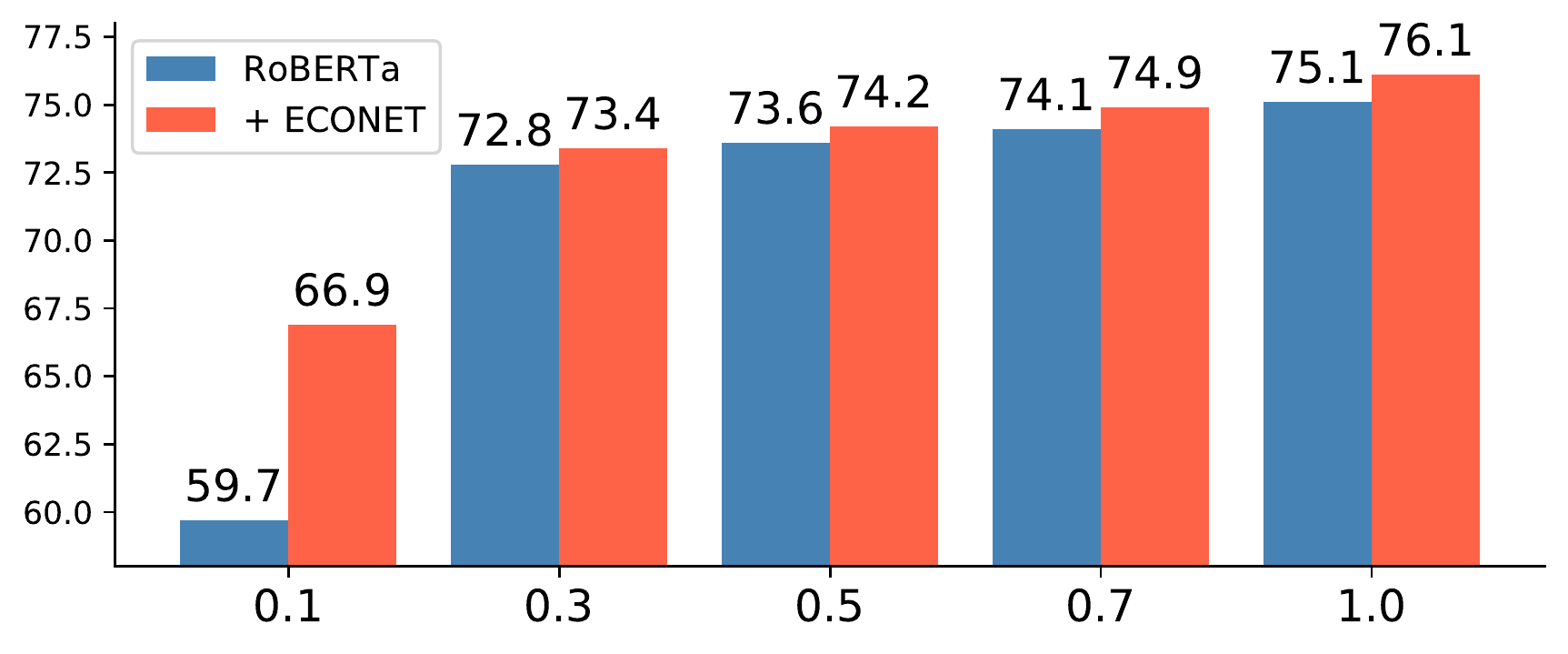}
    \caption{\torque}
    \label{fig:torque-ratio}
    \end{subfigure}
    \begin{subfigure}[b]{\columnwidth}
    \includegraphics[trim=0cm 0cm 0cm 0cm, clip, width=0.9\columnwidth]{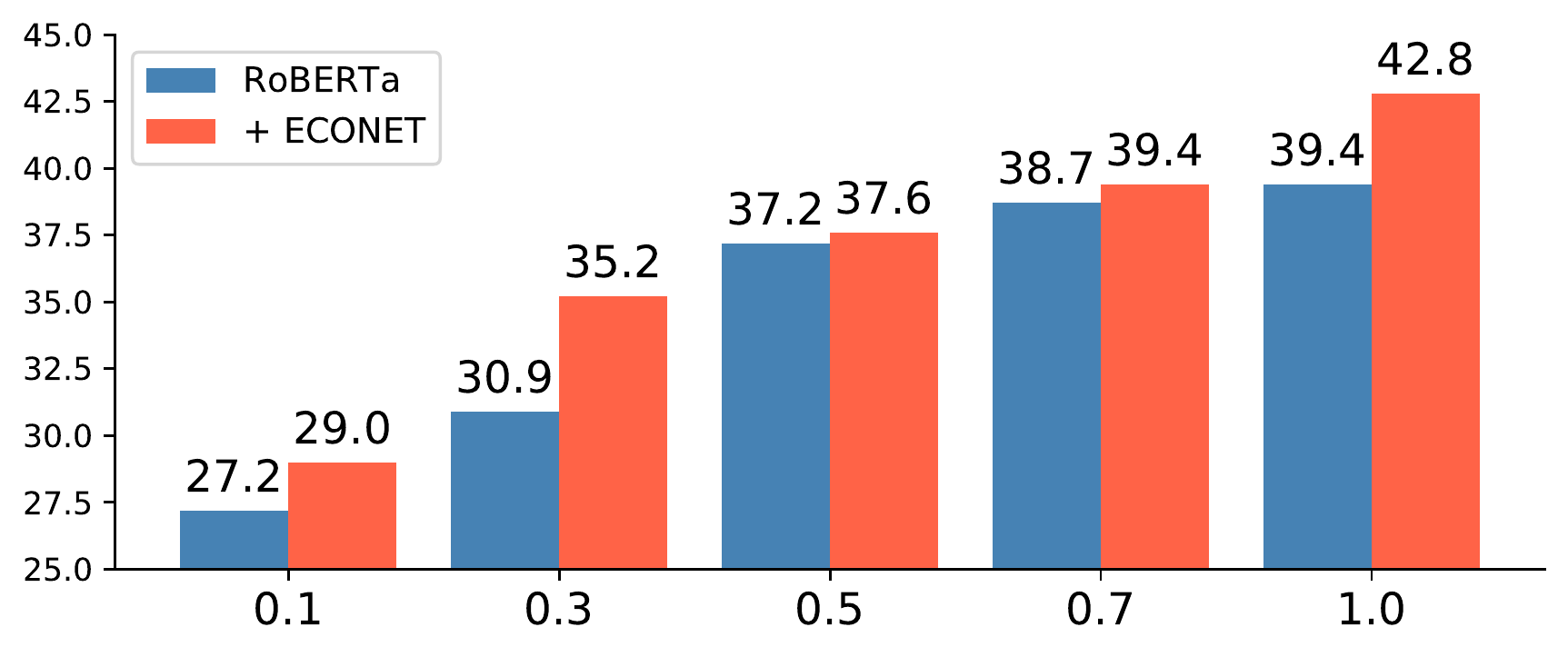}
    \caption{\red}
    \label{fig:red-ratio}
    \end{subfigure}
\vspace{-0.7cm}
\caption{Performances (F$_1$ scores) comparison between fine-tuning {\roberta} vs. {\roberta} + {\econet} over different ratios of the training data.}
\label{fig:ratios} 
\end{figure}


\begin{table*}[htbp!]
\centering
\scalebox{0.9}{
\setlength{\tabcolsep}{3.7pt}
\begin{tabular}{l|ccc|c|c|c}
\toprule
 & \multicolumn{3}{c|}{\textbf{\torque}} &{\textbf{\tbd}} & {\textbf{\matres}} & {\textbf{\red}}\\
\cmidrule(lr){2-4}\cmidrule(lr){5-5}\cmidrule(lr){6-6}\cmidrule(lr){7-7}
\textbf{Methods} & F$_1$ & EM & C & F$_1$ & F$_1$ &  F$_1$ \\
\midrule
TacoLM & 65.5($\pm$0.8) & 36.2($\pm$1.6) & 22.7($\pm$1.4) & 56.9($\pm$0.7) & 75.1($\pm$0.8) & 40.7($\pm$0.3)\\
\midrule
\bert & 70.9($\pm$1.0) & 42.8($\pm$1.7) & 29.0($\pm$0.9) & 56.7($\pm$0.2) & 73.9($\pm$1.0) & 40.6($\pm$0.0)\\
+ {\econet} & 71.8($\pm$0.3) & 44.8($\pm$0.4) & 31.1($\pm$1.6) & 56.1($\pm$0.8) & 74.4($\pm$0.4) & 41.7($\pm$1.5) \\
\midrule
\roberta & 76.7($\pm$0.2) & 50.5($\pm$0.5) & 36.2($\pm$1.1) & 59.8($\pm$0.3) & 79.9($\pm$1.0) & 43.7($\pm$1.0) \\
+ Generator & 76.6($\pm$0.1) & 51.0($\pm$1.0) & 36.3($\pm$0.8) & 61.5($\pm$0.9) & 79.8($\pm$0.9)& 43.2($\pm$1.2)\\
+ {\econet} & 76.9($\pm0.4$) & 52.2($\pm$0.9) & 37.7($\pm0.4$) & 60.8($\pm$0.6) & 79.5($\pm$0.1) & 44.1($\pm$1.7) \\
\bottomrule
\end{tabular}
}
\caption{Average Dev Performances Corresponding to Table~\ref{tab:results}. Note that for {\mctaco}, we train on dev set and evaluate on the test set as mentioned in Section~\ref{sec:experiments}, so we do not report test performance again here.} 
\label{tab:results-dev}
\end{table*}

\section{Variants of {\econet}}
\label{sec:variant}
We also experimented with a variant of {\econet} by first pretraining {\roberta} + Generator for a few thousands steps and then continue to pretrain with {\econet}. However, this method leads worse finetuning results, which seems to contradict the suggestions in \citet{zhou2020pretraining} and \citet{Clark2020ELECTRA} that the generator needs to be first trained to obtain a good prediction distribution for the discriminator. We speculate that this is due to our temporal and event mask predictions being easier tasks than those in the previous work, which makes the ``warm-up steps'' for the generator not necessary.

\section{Impact of Event Models}
Table~\ref{tab:event-model-impact} compares results based on two event annotations.
\begin{table}[h]
\centering
\setlength{\tabcolsep}{3.0pt}
\vspace{-0.2cm}
\scalebox{0.7}{
\begin{tabular}{lccc|c|c}
\toprule
& \multicolumn{3}{c|}{\textbf{\torque}} & {\textbf{TB-D}} & {\textbf{\red}}\\
\cmidrule(lr){2-4}\cmidrule(lr){5-5}\cmidrule(lr){6-6}
\textbf{Event Annotations} & F$_1$ & EM & C & F$_1$ & F$_1$ \\
\midrule
\textbf{\torque} &  \textbf{76.1} & \textbf{51.6} & \textbf{36.8} & \underline{64.8} & \textbf{42.8} \\
\textbf{\tbd} &  \textbf{76.1} & 51.3 & 36.4 & \underline{\textbf{65.1}} & \underline{42.6} \\
\bottomrule
\end{tabular}}
\vspace{-0.2cm}
\caption{Fine-tuning performance comparisons using event detection models trained on {\torque} v.s. {\tbd} event annotations. All numbers are average over 3 random seeds. Std. Dev. $\ge 1\%$ is underlined.}
\label{tab:event-model-impact}
\end{table}

\section{Ablation Studies for BERT}
\begin{table}[t]
\centering
\setlength{\tabcolsep}{3.0pt}
\scalebox{0.7}{
\begin{tabular}{lccc|c|c}
\toprule
& \multicolumn{3}{c|}{\textbf{\torque}} & {\textbf{TB-D}} & {\textbf{\red}}\\
\cmidrule(lr){2-4}\cmidrule(lr){5-5}\cmidrule(lr){6-6}
\textbf{Methods} & F$_1$ & EM & C & F$_1$ & F$_1$ \\
\midrule
\textbf{{\bert}} & \underline{70.6} & \underline{43.7} & \underline{27.5} & \underline{62.8} & 39.4\\
\textbf{+ random mask} & 70.6 & 44.1 & 27.2 & 63.4 & \underline{35.3}\\
\textbf{+ {\econet}} &  71.4 & 44.8 & 28.5 & 63.0 & 40.2 \\
\bottomrule
\end{tabular}}
\vspace{-0.2cm}
\caption{Fine-tuning performances. All numbers are average over 3 random seeds. Std. Dev. $\ge 1\%$ is underlined.}
\label{tab:abl-result-bert}
\end{table}

\section{Attention Scores}
\begin{figure}[h!]
\centering
    \begin{subfigure}[b]{\columnwidth}
    \includegraphics[trim=0cm 0cm 0cm 0cm, clip, width=0.95\columnwidth]{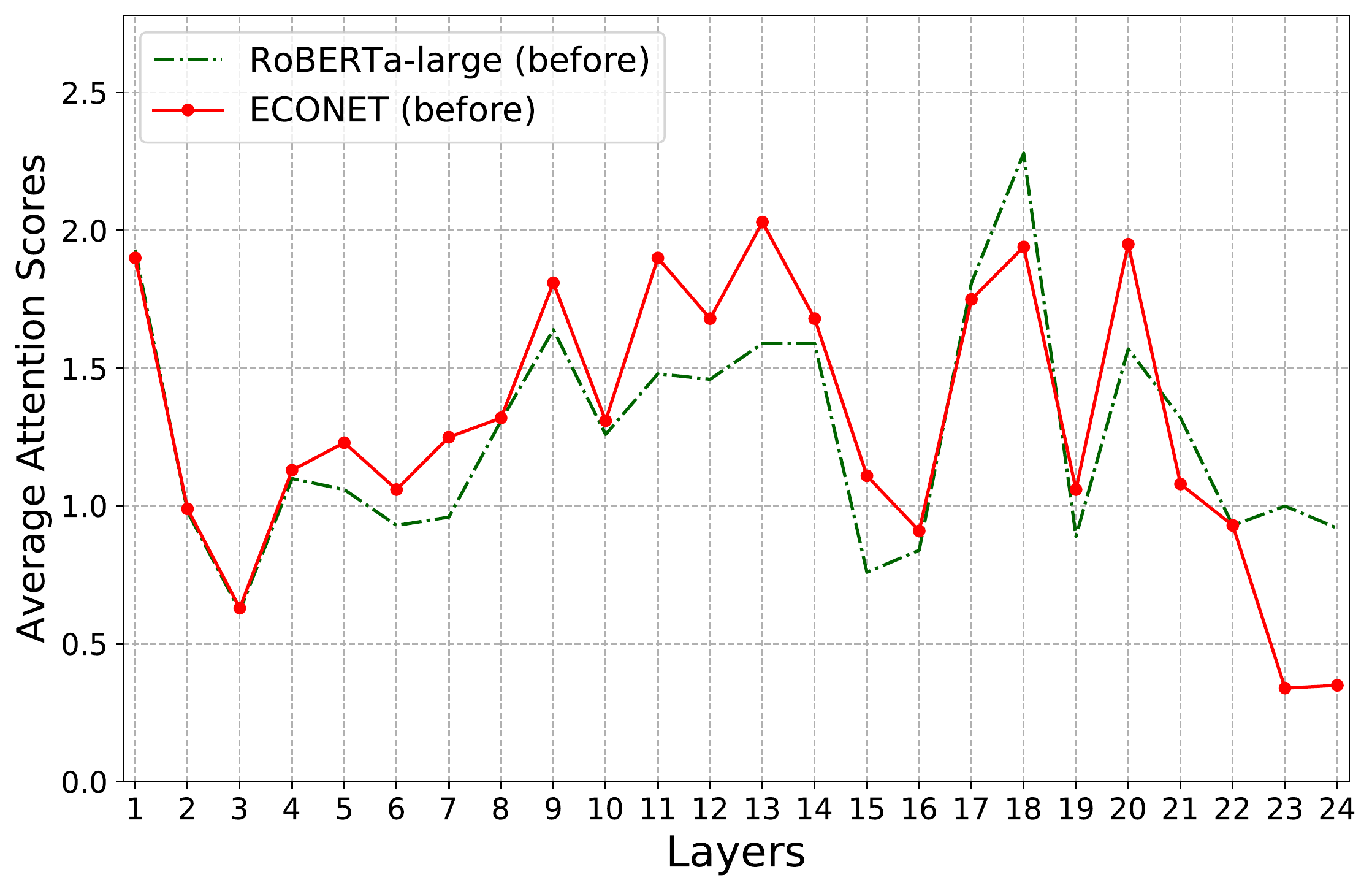}
    \vspace{-0.2cm}
    \caption{Attentions scores for [before] indicators.}
    \label{fig:tbd-before}
    \end{subfigure}
    
    \begin{subfigure}[b]{\columnwidth}
    \includegraphics[trim=0cm 0cm 0cm 0cm, clip, width=0.95\columnwidth]{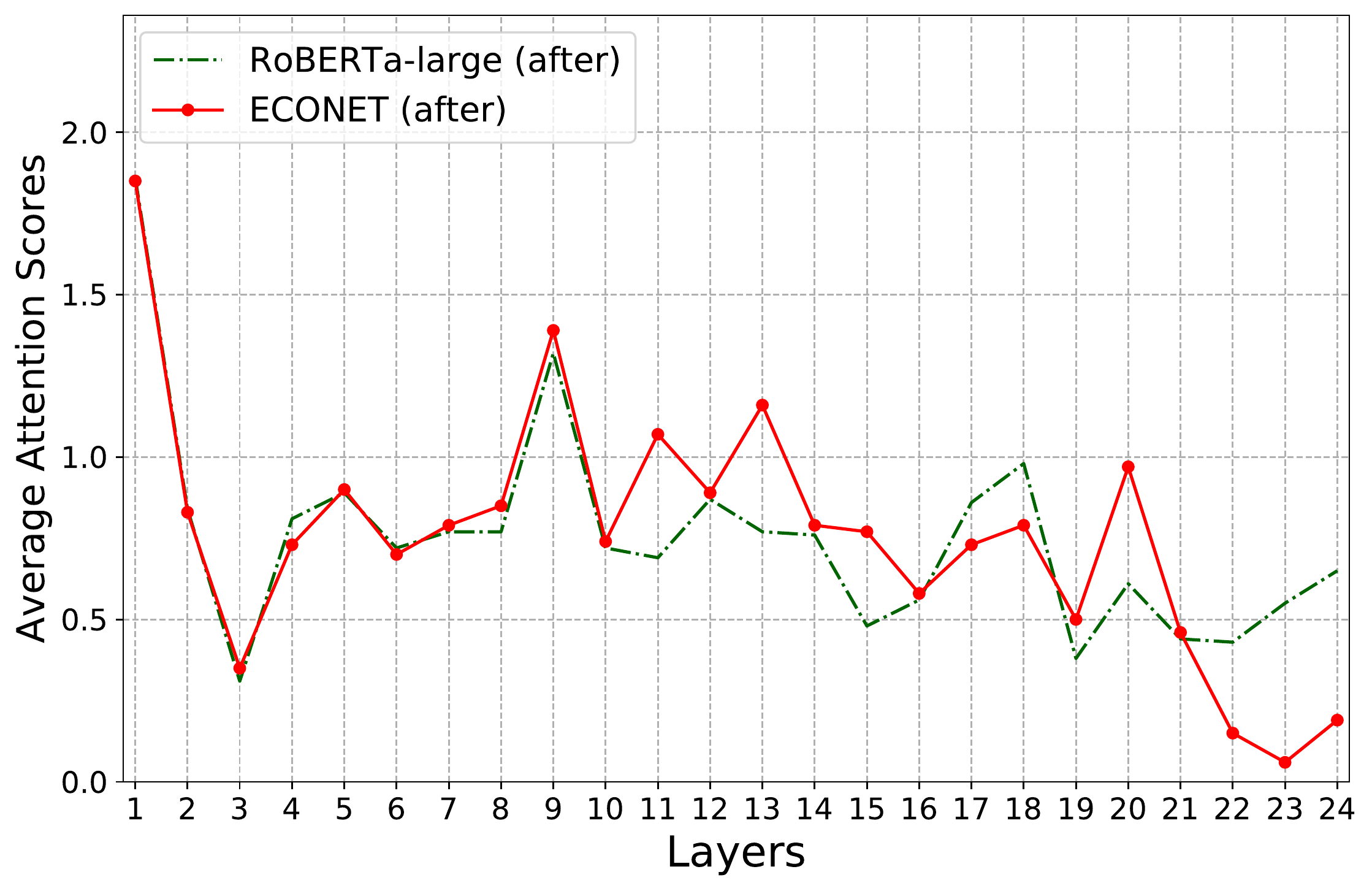}
    \vspace{-0.2cm}
    \caption{Attentions scores for [after] indicators.}
    \label{fig:tbd-after}
    \end{subfigure}
    
    \begin{subfigure}[b]{\columnwidth}
    \includegraphics[trim=0cm 0cm 0cm 0cm, clip, width=0.95\columnwidth]{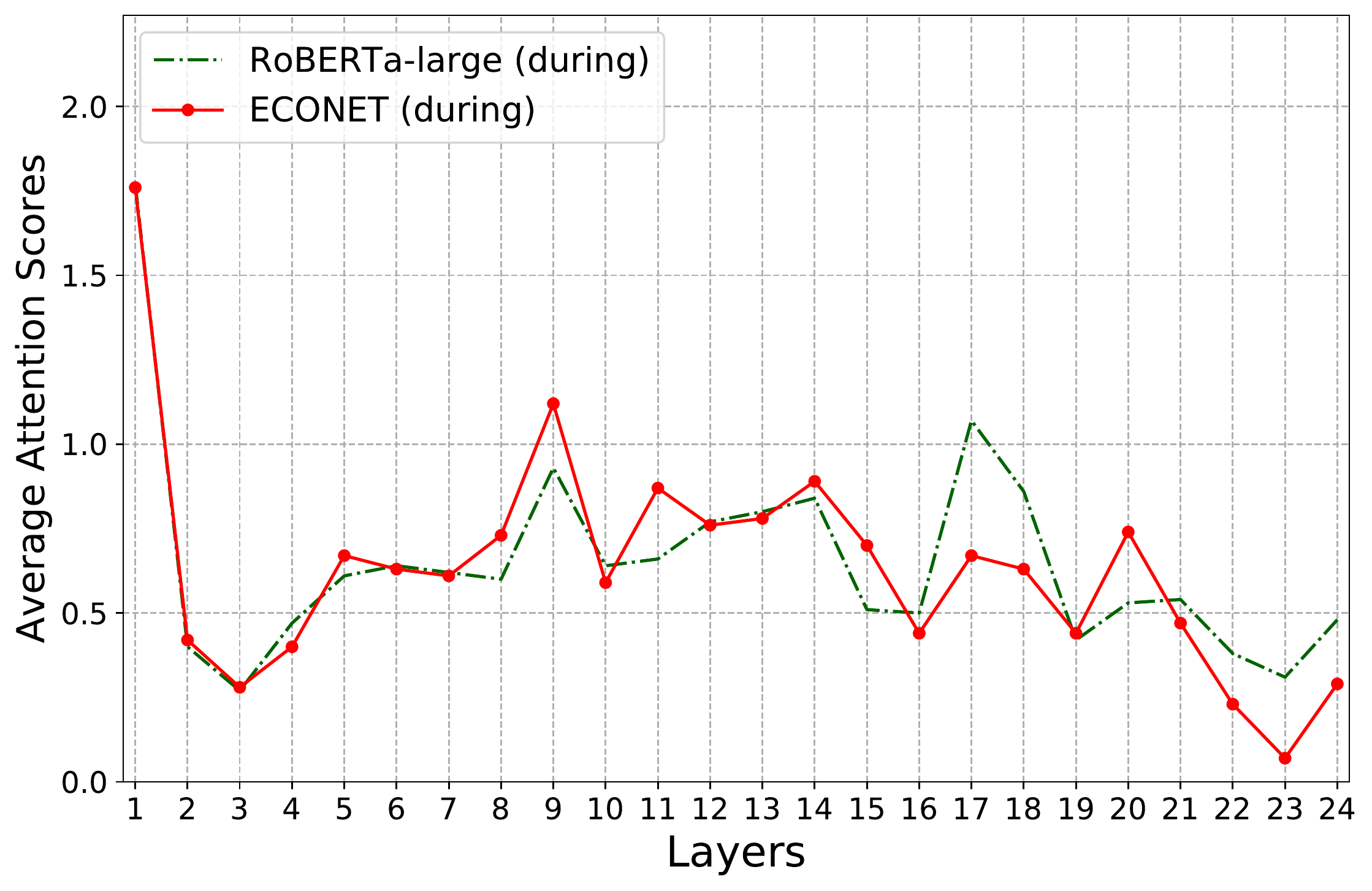}
    \vspace{-0.2cm}
    \caption{Attentions scores for [during] indicators.}
    \label{fig:tbd-during}
    \end{subfigure}

\caption{Attentions score comparisons between {\roberta} and {\econet} for all model layers. All numbers are multiplied by 100 and average over 3 random seeds for illustration purpose}
\label{fig:tbd-attn-all} 
\end{figure}

\section{Analysis of the Discriminator}
Table~\ref{tab:discriminator-res} shows the alignment of between the generator and the discriminator, and Table~\ref{tab:discriminator-ex} shows the examples of ``disagreed'' samples between the generator and the discriminator. Detailed analysis can be found in Section~\ref{sec:result} in the main text.

    
    

 \begin{table}[h!]
    \small 
 	\centering
 	\begin{tabular}{l|c|c|c|c}
 	\toprule
 	& \multicolumn{2}{c|}{Temporal Generator} & \multicolumn{2}{c}{Event Generator} \\
 	\cmidrule(lr){2-3}\cmidrule(lr){4-5}
 	& Corr. & Incorr. & Corr. & Incorr. \\
 	\cmidrule(lr){2-2}\cmidrule(lr){3-3}\cmidrule(lr){4-4}\cmidrule(lr){5-5}
 	Total \# & 837 & 163 & 26 & 974 \\
    \cmidrule(lr){1-1}\cmidrule(lr){2-2}\cmidrule(lr){3-3}\cmidrule(lr){4-4}\cmidrule(lr){5-5}
    Discr. Corr. \# & 816 & 29 & 25 & 964\\
    \cmidrule(lr){1-1}\cmidrule(lr){2-2}\cmidrule(lr){3-3}\cmidrule(lr){4-4}\cmidrule(lr){5-5}
    Accuracy & 97.5\%& 17.8\% & 96.2\%& 99.0\%\\
   \bottomrule
 	\end{tabular}
   	\caption{Discriminator's alignment with generator's mask predictions in {\econet}. Second column shows that discriminator strongly disagree with the ``errors'' made by the temporal generator.}
   	\label{tab:discriminator-res}
\end{table}

\begin{table}[h!]
    \small 
 	\centering
 	\begin{tabular}{l}
 	\toprule
 	\textbf{Type I. Same Group: 12/50 (24\%)}  \\ 
 	\midrule
 	\textbf{$\rangle$ $\rangle$Ex 1.} \textbf{original:} when; \textbf{predicted:} while \\
 	\midrule
 	\cellcolor[gray]{0.9}\textbf{Text:} A letter also went home a week ago in Pelham, in\\
 	\cellcolor[gray]{0.9}Westchester County, New York, \textbf{$\langle$mask$\rangle$} a threat made \\
 	\cellcolor[gray]{0.9}by a student in a neighboring town circulated in \\
 	\cellcolor[gray]{0.9}several communities within hours... \\
 	\midrule
 	 \textbf{$\rangle$ $\rangle$Ex 2.} \textbf{original:} prior to; \textbf{predicted:} before \\
 	 \midrule
 	\cellcolor[gray]{0.9}\textbf{Text:} ... An investigation revealed that rock gauges were \\ 
 	\cellcolor[gray]{0.9}picking up swifter rates of salt movement in the ceiling \\ 
 	\cellcolor[gray]{0.9}of the room, but at Wipp no one had read the computer \\
 	\cellcolor[gray]{0.9}printouts for at least one month \textbf{$\langle$mask$\rangle$} the collapse. \\
 	\midrule
    \textbf{Type II. Related Group: 8/50 (16\%)} \\
    \midrule
     \textbf{$\rangle$ $\rangle$Ex 3.} \textbf{original:} in the past; \textbf{predicted:} before \\\midrule
 	\cellcolor[gray]{0.9}\textbf{text:} Mr. Douglen confessed that Lautenberg, which \\
 	\cellcolor[gray]{0.9}had won \textbf{$\langle$mask$\rangle$}, was ``a seasoned roach and was \\ 
 	\cellcolor[gray]{0.9}ready for this race... \\
 	\midrule
 	 \textbf{$\rangle$ $\rangle$Ex 4.} \textbf{original:} previously; \textbf{predicted:} once \\
 	 \midrule
 	\cellcolor[gray]{0.9}\textbf{text:} Under the new legislation enacted by Parliament, \\ 
 	\cellcolor[gray]{0.9}divers who \textbf{$\langle$mask$\rangle$} had access to only 620 miles of the \\ 
 	\cellcolor[gray]{0.9}10,000 miles of Greek coast line will be able to explore \\
 	\cellcolor[gray]{0.9}ships and ``archaeological parks'' freely... \\
   \bottomrule
 	\end{tabular}
   	\caption{Categories and examples of highly related ``incorrect'' temporal indicator predictions by the generator, but labeled as ``correct'' by the discriminator.}
   	\label{tab:discriminator-ex}
 \end{table}
\end{document}